\pdfoutput=1

\documentclass[11pt]{article}

\usepackage{acl}

\usepackage{times}
\usepackage{latexsym}

\usepackage[T1]{fontenc}

\usepackage[utf8]{inputenc}

\usepackage{microtype}

\usepackage{inconsolata}

\usepackage{graphicx}

\usepackage{booktabs}
\usepackage{amsmath}
\usepackage{amssymb}
\usepackage[nameinlink,capitalize,noabbrev]{cleveref}
\usepackage[frozencache,cachedir=.]{minted}
\usepackage{nicefrac}
\usepackage{todonotes}
\usepackage{tikz}
\newcommand*\circled[1]{\tikz[baseline=(char.base)]{
            \node[shape=circle,draw,inner sep=1.5pt] (char) {\footnotesize#1};}}

\usepackage{enumitem}
\usepackage{multirow}
\usepackage{hyperref}

\title{GPT or BERT: why not both?}

\author{Lucas Georges Gabriel Charpentier$^*$ \and David Samuel$^*$ \\
  University of Oslo, Language Technology Group \\
  {\tt \{lgcharpe, davisamu\}@ifi.uio.no} \\
}

\begin{document}
\maketitle
\def\thefootnote{*}\footnotetext{Both authors contributed equally to this work.}\def\thefootnote{\arabic{footnote}}
\begin{abstract}
We present a simple way to merge masked language modeling with causal language modeling. This hybrid training objective results in a model that combines the strengths of both modeling paradigms within a single transformer stack -- \textsc{GPT-BERT} can be transparently used like any standard causal or masked language model. We test the pretraining process that enables this flexible behavior on the BabyLM Challenge 2024. The results show that the hybrid pretraining outperforms masked-only or causal-only models. We openly release the models, training corpora and code.\footnote{The models are available on HuggingFace at \href{https://huggingface.co/ltg/gpt-bert-babylm-base}{\texttt{ltg\-/gpt\--bert\--baby\-lm\--base}} and \href{https://huggingface.co/ltg/gpt-bert-babylm-small}{\texttt{ltg\-/gpt\--bert\--baby\-lm\--small}}; the corpora at \href{https://huggingface.co/datasets/ltg/babylm-2024-baby-cosmo-fine-100m}{\texttt{ltg\-/baby\-lm\--2024-ba\-by-cos\-mo-fine-100m}} and \href{https://huggingface.co/datasets/ltg/babylm-2024-baby-cosmo-fine-10m}{\texttt{ltg\-/baby\-lm\--2024-ba\-by-cos\-mo-fine-10m}}. The training scripts are available on GitHub at \href{https://github.com/ltgoslo/gpt-bert}{\texttt{ltg\-oslo\-/gpt\--bert}}}
\vspace{1em}
\end{abstract}

\section{Introduction}
\label{sec:introduction}

Language models have become fundamental tools in natural language processing, with two dominant paradigms: causal language models (CLM) and masked language models (MLM). Six years ago, GPT by \newcite{Radford2018ImprovingLU} demonstrated the generative abilities of transformer-based causal language models. Just a few months after this publication, BERT by \newcite{devlin-etal-2019-bert} heavily outperformed the causal GPT models when finetuned on downstream NLP tasks, showcasing the major advantage of masked language modeling. These two `historical' models define the main use-cases of the two paradigms up to this date.

The difference between these paradigms lies in how they process text. CLMs can only look at previous tokens when making predictions, mimicking the left-to-right reading process. This makes them particularly well-suited for efficient text generation. MLMs, on the other hand, can access both previous and following tokens, allowing them to build richer contextual representations. This bidirectional context has proven especially valuable for tasks requiring deep language understanding.

\begin{figure}[!t]
        \centering
        \includegraphics[width=\linewidth]{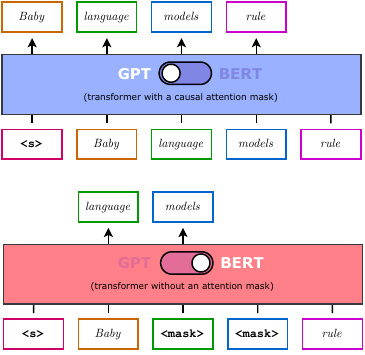}
        \caption{\textbf{Two modes of a single model}\hspace{1.5em}Causal and masked language modeling can be easily unified by shifting both outputs by one token to the right. Then we can train one language model on both paradigms at the same time just by modifying the input tokens, output tokens and attention masks.}
        \label{fig:hybrid}
    \end{figure}

\paragraph{BERTs should not be forgotten}
A recent paper by \newcite{samuel2024berts} revealed that BERT-like model are just as capable text generators as GPT-like models. Yet, when these two types of models are evaluated on a shared ground -- generative in-context learning \citep{NEURIPS2020_1457c0d6} -- they still show radical differences, clearly outperforming each other in different areas. Each paradigm has its own strengths and combining them into a single hybrid might lead to a model with a more general language understanding.

\paragraph{GPT-BERT}
This motivated us to introduce \textsc{GPT-BERT}, a hybrid language model that combines the strengths of both CLM and MLM approaches. Our key insight is that the two objectives can be unified by reformulating how output tokens are handled in the MLM framework. Instead of predicting masked tokens at their original positions, we shift the predictions one position to the right, aligning them with the CLM's next-token prediction pattern. This simple modification allows us to train a single model that can seamlessly switch between masked and causal modes without any architectural changes or additional parameters.

This paper demonstrates the benefits of the hybrid approach across multiple benchmarks. We evaluate \textsc{GPT-BERT} on the \textit{BabyLM Challenge 2024} \citep{babylm-2024}, which provides a controlled environment for comparing language models trained on limited data. Additionally, we explore the impact of varying the ratio between MLM and CLM, and we test the model's ability to perform in-context learning and text generation.

The results suggest that integrating MLM and CLM objectives during pretraining leads to more robust and capable language models, even in low-resource scenarios, without any extra training cost. Our approach opens up new possibilities for developing more efficient and versatile models for a wide range of natural language processing tasks.

\section{Method}

\subsection{Hybrid masked-causal language modeling}
\label{sec:hybrid}

    In order to align both objectives we use a slightly modified version of masked language modeling called \textbf{masked next-token prediction} \citep[MNTP;][]{behnamghader2024llmvec}. The only difference from traditional MLM is that when the token at position $k+1$ is masked, its prediction should be outputed at position $k$. In this way both MLM and CLM are unified as the output at position $k$ always represents the token at position $k+1$. These two modes are illustrated in \Cref{fig:hybrid}.
    
    \paragraph{Dataset handling} To ensure that our model sees all the data for both objectives, we duplicate our dataset. One is used for the causal objective, and the other for the masked objective. We can then decide a ratio of causal-to-masked in which to divide the data seen by the model at each batch.
    
    \paragraph{Loss and transformer architecture} No additional changes are needed. Both training objectives minimize the cross-entropy loss, they share all learnable parameters, and use the same transformer encoder/decoder module.

\subsection{Other modifications}
\label{sec:modification}

We base the transformer architecture of our models on LTG-BERT \citep{samuel-etal-2023-trained}, but make some additional modifications to improve its performance. These changes are ablated in \Cref{sec:experiments}.

\paragraph{Attention gate}

Following \newcite{AlphaFold2021}, we gate the outputs of the attention operation. This is akin to the gated linear units (GLU) that have been proposed to improve the expressivity of feed-forward modules \citep{shazeer2020gluvariantsimprovetransformer}. This modification also simplifies the definition of the transformer architectures, now both the attention modules and the feed-forward modules can be expressed as:

\begin{minted}[linenos=false, breaklines=true, baselinestretch=1.2, breakanywhere=true, fontfamily=tt, fontsize=\footnotesize, numbersep=12pt, xleftmargin=0.0em,firstnumber=1,escapeinside=@@]{python}
def layer(x: @\textbf{\texttt{tensor}}@, layer_id: int):
    residual = x            # skip-connection
    x = layer_norm(x)       # without parameters
    g = gate(x)             # linear projection
    if layer_id % 2 == 0:   # if attention layer
        x = attention(x)    # do attention
    else:                   # else feed-forward
        x = linear(x)       # linear projection
    x = glu(x, g)           # activation (GEGLU)
    x = layer_norm(x)       # without parameters
    x = output(x)           # linear projection
    return residual + x
\end{minted}


\paragraph{Layer weighting} We further increase the expressivity of the transformer backbone by allowing each layer to select its desired combination of outputs from previous layers. This directly follows the ELC-BERT models \citep{georges-gabriel-charpentier-samuel-2023-layers} and the later modification by \newcite{pagliardini2024denseformer} who allow any linear combination of layers instead of restricting the combination to be convex. We also make the weighting more granular by treating the attention and feed-forward modules as separate layers. With each $\alpha_{ij} \in \mathbb{R}$ being a learnable scalar, the forward pass of the resulting transformer works as follows:

\begin{minted}[linenos=false, breaklines=true, baselinestretch=1.2, breakanywhere=true, fontfamily=tt, fontsize=\footnotesize, numbersep=12pt, xleftmargin=0.0em,firstnumber=1,escapeinside=@@]{python}
def transformer(subword_indices: @\textbf{\texttt{tensor}}@):
    output@$_{\texttt{0}}$@ = embedding(subword_indices)
    for i in range(1, n_layers + 1):
        output@$_{\texttt{i}}$@ = @$\sum_{\texttt{j=1}}^{\texttt{i}}{\alpha_{\texttt{ij}}\cdot\texttt{layer}(\texttt{output}_{\texttt{j-1}}, \texttt{j})}$@
    return output@$_{\texttt{n\_layers}}$@ 
\end{minted}


\paragraph{Batch-size scheduling} We improve the sample-efficiency (and speed) of pretraining by linearly increasing the batch size during training \citep{rae2022scalinglanguagemodelsmethods, deepseekv2}. The intuition behind this method is that high-quality gradients are mainly needed at the late stages of pretraining, the initial steps can be guided by good-enough gradients from smaller batches. The maximum batch size is taken from LTG-BERT (4M tokens), but we start the training with just $\nicefrac{1}{4}$ of this value, thus dividing the total number of tokens needed for training by $2$.

\paragraph{Mask scheduling} Another way to increase the sample-efficiency is to recover more unmasked tokens during training. However, \newcite{ankner-etal-2024-dynamic} showed that this might be in conflict with the downstream usage of MLMs. Thus they propose to linearly decrease the masking probability throughout the training, starting with $30\%$ and finishing with the standard $15\%$ masking. We adopt this scheme, believing that it also reduces the impact of smaller batches at the beginning of training.

\section{Pretraining and evaluation}

The main purpose of this section is to evaluate if the MLM and CLM training objectives can be merged, and to evaluate the effect of this. We base the experiments on the BabyLM challenge \citep{babylm-2024}.

\paragraph{BabyLM challenge}

This shared task provides a shared ground for experiments on small-scale language modeling. Its second iteration consists of four tracks: \textsc{strict}, \textsc{strict-small}, \textsc{vision} and \textsc{paper}. We participate in the first two text-based tracks. There, the submissions have to be pretrained solely on a fixed number of words, 100M in the \textsc{strict} track and about 10M words in the \textsc{strict-small} track. The organizers do provide a default dataset for each track, but unlike the previous edition, the participants are not limited to using it, as long as they stay under the word count limit. For the \textsc{vision} track, the participants are limited to 50M words and as many images as they want. Here the goal is to create a multi-modal model. Finally, the \textsc{paper} does not require the submission of a model to the task. This track encourages contributions related to the goal of the challenge such as new cognitively-inspired metrics. As detailed in \cref{sec:evaluation}, the submissions are compared on a shared evaluation set consisting of syntactic and natural language understanding tasks.

\begin{table}[t!]
\resizebox{\columnwidth}{!}{%
\begin{tabular}{@{}l@{\hspace{-1em}}rrrr@{}}
{\small\textsc{strict-small} track (10M words)}\\
\toprule
\textbf{Model} & \textbf{BLiMP $\uparrow$} & \textbf{BLiMP-S $\uparrow$} & \textbf{GLUE $\uparrow$} & \textbf{EWOK $\uparrow$}\\\midrule
Encoder-only \textsubscript{\textit{(BabyLM baseline)}} & 60.6 & 60.8 & 60.3 & 48.9 \\
Decoder-only \textsubscript{\textit{(BabyLM baseline)}} & 69.8 & 59.5 & 63.3 & 50.7  \\[0.75em]
ELC-BERT \textsubscript{\textit{(2023)}} & 80.5	& 67.9 & 75.3 & 51.0 \\
LTG-BERT \textsubscript{\textit{(2023)}} & 80.6 & \textbf{69.8} & 74.5 & ---\\[0.75em]
GPT-BERT \textsubscript{\textit{(ours)}} & \textbf{81.2} & 69.4 & \textbf{76.5} & \textbf{54.6} \\
\bottomrule
\\
{\small\textsc{strict} track (100M words)}\\
\toprule
\textbf{Model} & \textbf{BLiMP $\uparrow$} & \textbf{BLiMP-S $\uparrow$} & \textbf{GLUE $\uparrow$} & \textbf{EWOK $\uparrow$}\\\midrule
Encoder-only \textsubscript{\textit{(BabyLM baseline)}} & 69.2 & 66.5 & 68.4 & 51.9 \\
Decoder-only \textsubscript{\textit{(BabyLM baseline)}} & 73.1 & 60.6 & 69.0 & 52.1 \\[0.75em]
ELC-BERT \textsubscript{\textit{(2023)}} & 85.8	& \textbf{76.8}  & 78.3 & 56.3 \\
LTG-BERT \textsubscript{\textit{(2023)}} & 85.3 & 76.6  & 77.9 & 56.0 \\[0.75em]
GPT-BERT \textsubscript{\textit{(ours)}} & \textbf{86.1} & \textbf{76.8} & \textbf{81.5} & \textbf{58.4} \\
\bottomrule
\end{tabular}%
}
\caption{\textbf{BabyLM submission scores}\hspace{1.5em}The final scores of our \textsc{strict-small} and \textsc{strict} models submitted to the BabyLM challenge \citep{babylm-2024}. The table also includes the winner of the last year's iteration of this shared task (ELC-BERT), the baseline for our current model (LTG-BERT), as well as the baselines provided by the organizers. Results of other submission were not available as of writing this paper. Higher scores are better, the best results in each evaluation suite are boldfaced.
}
\label{tab:babylm-scores}
\end{table}

\paragraph{Training corpus} We pretrain both submissions on a $1:1:1$ mix of the provided BabyLM corpus, on a subset of the FineWeb-Edu corpus \citep{lozhkov2024fineweb-edu}, and on a small subset of the Cosmopedia corpus \citep{benallal2024cosmopedia}. The main purpose of training on this mixture is to provide the model with more factual knowledge and more diverse language. 

\paragraph{Pretraining process} Generally speaking, we adopt the training recipe of LTG-BERT \citep{samuel-etal-2023-trained}, which was optimized for pretraining on another low-resource 100 million English corpus.\footnote{\url{https://github.com/ltgoslo/ltg-bert}} The pretraining process is the same for both tracks, except for using a smaller vocabulary and a smaller model for the \textsc{strict-small} track.

As for the \textsc{strict} track, we use a \textsc{base}-sized language model with 119 million parameters. We train a case-sensitive BPE tokenizer \citep{Gage1994ANA} with a vocabulary size of $2^{14} = 16\,384$, using solely texts from the training corpus. The \textsc{base} is trained for 15\,625 steps with an average batch size of 2 million tokens. The \textsc{strict-small} track is tackled by a \textsc{small}-sized language model with 30 million learnable parameters. The subword vocabulary is reduced to $2^{12} = 8\,192$ items. The training steps of the \textsc{small} model are reduced to 7\,812. The full list of hyperparameters and implementation details are provided in \cref{app:training}.

\paragraph{Evaluation}
\label{sec:evaluation}

We utilize the language modeling benchmark suite from the BabyLM challenge \citep{eval-harness, babylm-2024},\footnote{\url{https://github.com/babylm/evaluation-pipeline-2024}} which relies on three conceptually different evaluation tasks:
\begin{enumerate}
    \item The GLUE and SuperGLUE datasets test the ability of a pretrained model to adapt to various language understanding tasks.
    \item BLiMP and BLiMP-supplement tasks test the affinity of a model towards grammatical sentences in a completely zero-shot manner.
    \item EWOK is another zero-shot task. It tests the ability of a model to understand concepts such as spatial relations or physical dynamics.
\end{enumerate}

\noindent
We further elaborate on each of these evaluation suites in \Cref{app:evaluation}.

\section{Experiments}
\label{sec:experiments}

\begin{figure*}[!h]
    \centering
    \includegraphics[width=\textwidth]{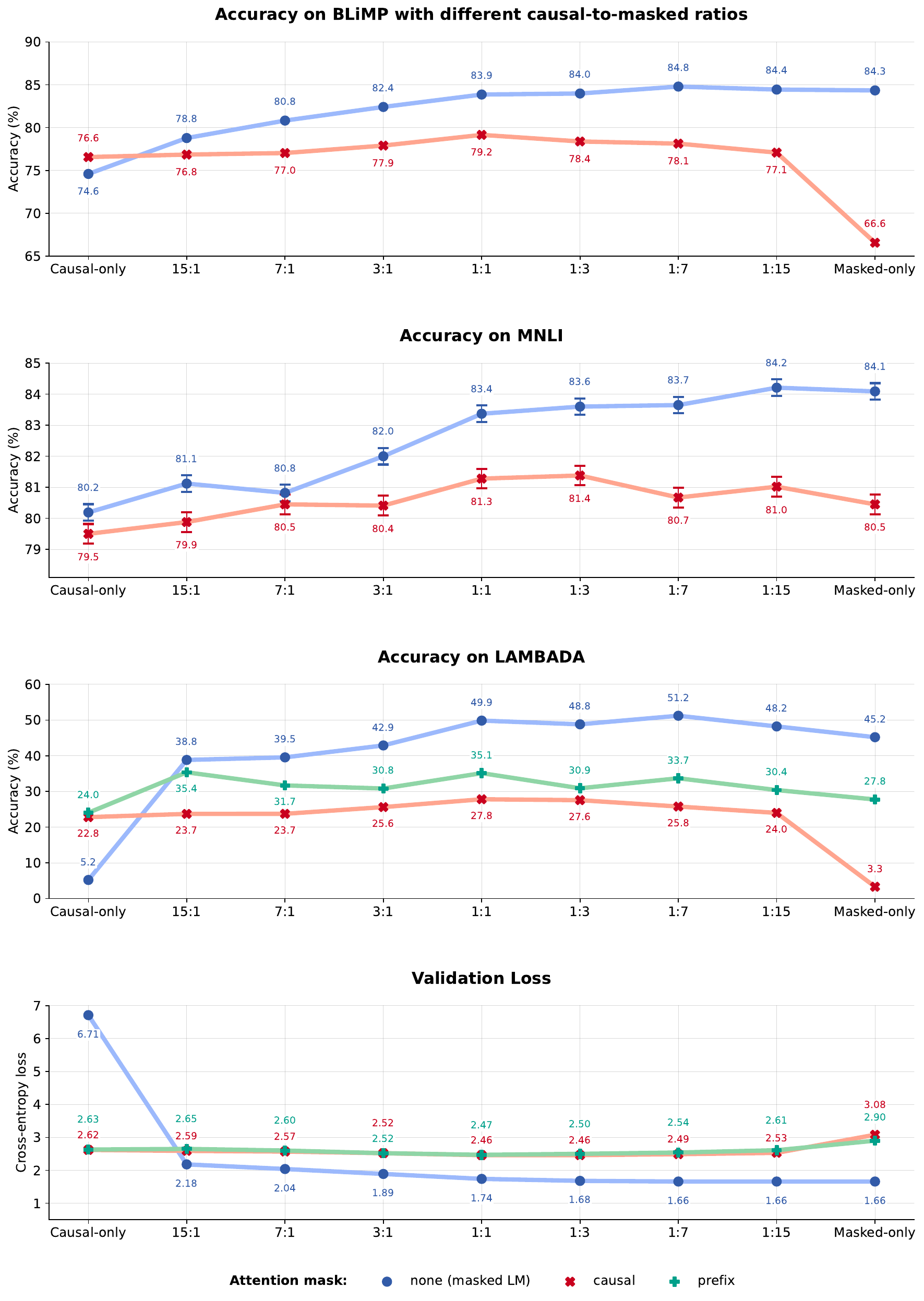}
    \caption{\textbf{The effect of the causal-to-mask ratio}\hspace{1.5em}Comparison of performance of different tasks when varying the ratio of MNTP used during pre-training. We also look at the performance of the model using prefix language modeling with a partially-bidirectional attention mask. MNLI scores are reported with standard deviation error bars estimated by averaging the variations across three finetuning random seeds.}
    \label{fig:lambada}
\end{figure*}

\subsection{BabyLM submission}

\Cref{tab:babylm-scores} shows the performance of our models against the backbone architecture of the model (LTG-BERT), as well as last year's winner on both tracks (ELC-BERT). 
We can see that for the \textsc{strict-small} track our model outperforms last year's winner in every benchmark and is only beaten by LTG-BERT on BLiMP-Supplement by 0.4. For our submission to the \textsc{strict} track our model outperforms or matches both models (only ELC-BERT on BLiMP-Supplement matches our model). One thing to note, is that the filtration of the evaluation datasets are slightly different leading to comparisons between not exact.

For completeness, in \Cref{tab:babylm-scores}, we also include the performance of the models provided by the BabyLM organizers  \citep{babylm-2024}. The provided encoder-only models are based on LTG-BERT \citep{samuel-etal-2023-trained}, and the decoder-only models are based on Baby Llama \citep{timiryasov-tastet-2023-baby}. Our models clearly outperforms these baselines on all metrics, but that might be mostly attributed to their smaller pretraining budget.

\subsection{Masked or causal?}

Since our model can learn both from masked and causal examples, the question becomes, whether using a combination of both is better than using only one of the two methods during pretraining. To evaluate this, we look at the performance of models pretrained with different causal-to-masked ratios. 

The main results are presented in \Cref{fig:lambada}. We evaluate the models on four tasks that cover distinct uses: \circled{1} BLiMP is a zero-shot linguistic-preference task that is typically better suited for masked language models \citep{salazar-etal-2020-masked}; \circled{2} MNLI is a popular dataset for evaluating the finetunability of a language model, which also benefits masked language models; \circled{3} LAMBADA, on the other hand, is a language modeling dataset mostly used to evaluate causal language models; and \circled{4} we also directly compute the validation loss of each model. Furthermore, when applicable, each task is tested with three settings: fully-bidirectional processing (without any attention mask), unidirectional processing (with a causal mask), and partially-bidirectional processing (with a prefix mask).

The validation loss of the causal and prefix masking is calculated on the second half of the tokens of a given input sequence, where the first half of the tokens are either seen in a bidirectional fashion (prefix) or in a causal fashion (causal). For LAMDABA the entire context is seen bidirectionally for the prefix evaluation. Finally, when fine-tuning MNLI with the causal mask, we use the same tokenization as \citet{Radford2018ImprovingLU} where a both a delimiter token is added in-between the two sentences as well as a extract token at the end of the input (two different tokens are used).

For the MNLI hyperparameters, we did a sweep on the SST-2 dataset for each model and took the best performing hyperparameters for each model and each masking (i.e. each model and masking scheme had their own hyperparameters). We sweeped over $\{1, 3, 5\}$ for number of epochs, $\{3\cdot10^{-5}, 5\cdot10^{-5}, 1\cdot10^{-4}\}$ for learning rates, and $\{16, 32\}$ for batch sizes.

\paragraph{Bidirectional results} If we start by focusing on the bidirectional results, we see that the best results for all the tasks can be found for the models with a lower causal-to-masked ratio (from 1:7 to masked-only). More specifically, the 1:7 model is the best on BLiMP and LAMBADA, the best model for MNLI is 15:16, and both those models and the masked-only model achieve the best results on the validation loss. We also see that adding as few as $6.25\%$ MNTP training can lead to significant increases in bidirectional performances ($+4.2\%$ on BLiMP, $+0.9\%$ on MNLI, $+33.3\%$ on LAMBADA and $-4.53$ on validation loss). In addition, using a bidirectional mask for evaluation performs the best for all models except the causal-only, however, this is unsurprising given this model is never trained to attend to every token.

\paragraph{Causal LM results} Looking at the results when using causal masking, we see that the best models shift towards a more balanced ratio between the causal and masked training objectives. The 1:1 model and 1:3 model perform roughly the same on all tasks. As mentioned before, the results are worse than for the bidirectional evaluation; most likely because of the lower expressivity of causally-masked models \citep{ewer2024entpencoderonlytokenprediction}. Further focusing on MNLI, we see that the purely causal model does not truly benefit from being finetuned with a bidirectional mask (only $+0.7\%$ improvement, with the results being within two standard deviations of each other). Once we add some MNTP training we see a significant difference in the results between both masking strategies. With only $6.25\%$ MNTP added, we have a $1.2\%$ improvement when using the bidirectional mask. This trend grows to being an over $3\%$ improvement in performance.

\paragraph{Prefix LM results} Finally, we look at the performance for the prefix masking (partially bidirectional). We only evaluate prefix masking on LAMBADA and validation loss since it would be difficult to do this for both BLiMP and MNLI. We see that on validation loss we get similar (if not slightly worst) results as for the causal masking while the results on LAMBADA are slightly improved. In addition, the LAMBADA results do not have a clear trend outside of the hybridized models performing better than the single-objective models. This leads us to believe that our models can perform limited prefix language modeling even though they were not explicitly trained to do so.


\paragraph{Other benchmarks} Similar trends can be seen on the other datasets in \cref{app:vary}. Based on the results on all tasks, we decided to use a 1:15 causal-to-masked ratio for our final model (to which every model is compared in subsequent sections) as well as the bidirectional evaluation scheme. In \cref{sec:icl,sec:generation}, a model trained on this ratio is used for the in-context learning and text generation.

\subsection{Ablation study}

\begin{table}[!t]
\resizebox{\columnwidth}{!}{%
\begin{tabular}{@{}l@{\hspace{1em}}rrrr@{}}
{\small\textsc{strict-small} track (10M words)}\\
\toprule
\textbf{Model configuration} & \textbf{PPL $\downarrow$} & \textbf{BLiMP $\uparrow$} & \textbf{MNLI $\uparrow$} & \textbf{EWOK $\uparrow$} \\ \midrule
GPT-BERT                   & \textbf{10.8} & \textbf{81.2} & 80.1 & 54.6 \\[0.75em]
\textit{without} layer weights    & $+$0.4          &$-$1.3 & $+$0.2 & $+$0.6 \\ 
\textit{without} attention gate   & $+$0.3          & $-$0.3 & $+$0.3 & $-$0.9 \\
\textit{without} mask scheduling  & $+$0.1          & $-$0.1 & $-$0.7 & $-$0.6  \\
\textit{without} batch scheduling & $+$0.7          &  $-$1.1 & 0.0 & \textbf{$+$0.8} \\[0.75em]
\textit{with only} BabyLM corpus & --- & $-$0.2 & $-$1.6 & $-$2.0\\
\textit{with only} FineWeb-edu   & --- & $-$0.4 & \textbf{$+$1.1} & $-$0.8\\
\textit{with only} Cosmopedia    & --- & $-$7.1 & 0.0 & $-$0.6 \\ \bottomrule
\end{tabular}%
}
\caption{\textbf{Ablation study}\hspace{1.5em}Comparison of different model configurations proposed in \Cref{sec:modification}, and corpus mixtures. The top row shows the performance of the final model (with all modifications), the middle rows show the absolute performance difference of models with one modification less, and the last group of rows shows the performance difference of GPT-BERT models trained on corpora from single sources.}
\label{tab:ablation}
\end{table}

\begin{figure*}[!th]
    \centering
    \includegraphics[width=\textwidth]{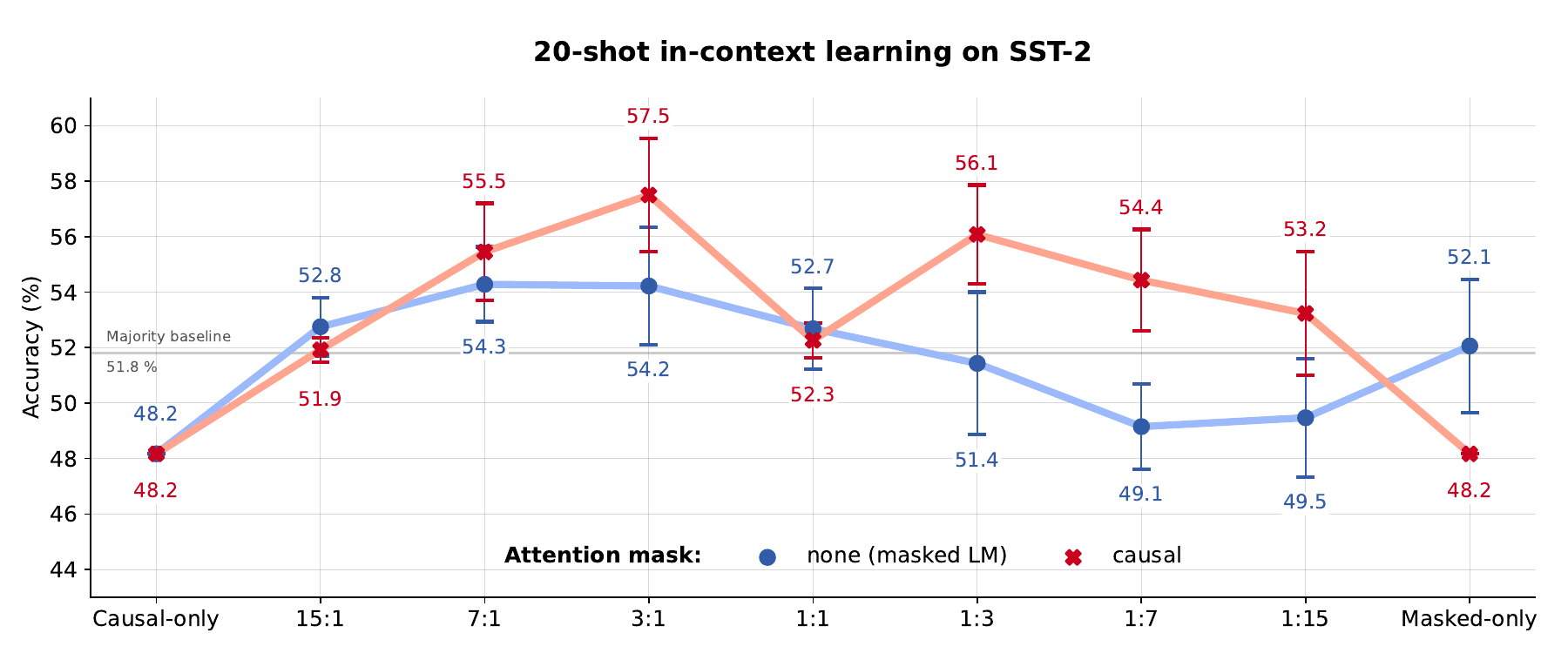}


    \caption{\textbf{SST-2 in-context learning}\hspace{1.5em}20-shots ICL results on the SST-2 validation set for models trained on the 100M BabyLM datasets with varying degrees of each objective. The demonstrations (shots) were chosen at random from the training dataset. We do 20-runs and report mean as well as standard deviation. Note that the accuracy of the majority baseline on this dataset is 51.8\%.}
    \label{tab:icl_sst2_varying}
\end{figure*}

We ablate the modeling choices from \Cref{sec:modification} as well as different choices of training data. We train the ablated models with the \textsc{strict-small} setup and evaluate them on BLiMP, EWOK and MNLI (the largest GLUE dataset). The ablation results are in \Cref{tab:ablation}.

\paragraph{Results of the transformer ablation} All our modeling decisions during development were based on the training and validation perplexities -- this ablation study therefore provides an informative comparison based on a `held-out' downstream performance. \circled{1} In particular, the value of learnable layer weights is not clear for GPT-BERT, especially considering that they substantially slowdown the training (almost $1.5\times$). \circled{2} Attention gating, on the other hand, seems to be a better substantiated improvement, which also does not add any major computational cost. \circled{3} Mask scheduling is definitely a recommended improvement for any BERT-like models, based on all scores in this study. \circled{4} Batch scheduling does not show on overall negative impact, which means that GPT-BERT can be trained $2.0\times$ more efficiently using this method without a noticable degradation.

\paragraph{Results of the corpus ablation} The ablation of the three text corpora used for training our submission shows how each of them excels in a different area -- the BabyLM collection outperforms the others on BLiMP, FineWeb-Edu significantly improves the MNLI performance, and Cosmopedia is the best corpus for learning EWOK-style factual knowledge. The performance of the final GPT-BERT models shows that taking a random $\nicefrac{1}{3}$ of each corpus and combining them into a single collection works surprisingly well; combining the strengths of each corpus.

\subsection{Text generation}
\label{sec:generation}

None of the BabyLM benchmarks is particularly well-suited for assessing CLM performance. Thus, apart from LAMBADA, we also look at the generation capabilities of \textsc{GPT-BERT}. First we try standard greedy decoding with no repetition penalty (omitting new lines for readability):

{\small
\begin{description}
    \item[\texttt{It was a calm night}]\texttt{in the small town of Harmonyville, and the residents were eager to spend the day exploring the beautiful park. Among them were two best friends, Sam and Alex, who loved to learn about nature and its wonders.}
    \item[\texttt{Brazil is known for}]\texttt{its rich history and culture. It has been home to many indigenous peoples, including the indigenous people of the Amazon rainforest. One of the most famous indigenous groups in Brazil is the Brazilian people. They are known for their strong culture, traditional beliefs, and unique traditions.}
\end{description}
}

\noindent
As we can see that model is relatively repetitive, especially in the second generation, where some words are repeated in almost every sentence. If we now give the second prompt with a repetition penalty of 1.5, we get:

{\small
\begin{description}
    \item[\texttt{Brazil is known for}] \texttt{its rich history and culture. It has been home to many indigenous peoples, including the Mayan civilization, who have built impressive cities like Tiapaca (present-day Uruguay), Cusco (now Guatemala), Chihuahua (also called Yucatán), Puebla (which was once part of Mexico), Huachimaso (then modern-day Colombia), and more. One of the most famous places in Brazil is the city of Rio de Janeiro, located on the southern tip of the Amazon basin.}
\end{description}
}

\noindent
Although the model is not factually correct, it stays on topic while generating meaningful and well-formed text.

Overall, our model seems to be able to generate text, even though it has a hard time remembering exact facts and stay on topic. However, without applying repetition penalty the model struggles with repeating itself. More generations using prompts from \citet{radford2019language} can be found in \cref{app:generation}.

\subsection{In-context learning}
\label{sec:icl}

A well-known ability of larger language models is to use in-context information given in prompts to solve tasks without any finetuning -- causal ones \citep{NEURIPS2020_1457c0d6}, as well as masked models \citep{samuel2024berts}. However, these capabilities are often thought to appear only once a model is large enough or trained on a vast amount of data \citep{wei2022emergent}. 

Despite the number of parameters and the size of the training corpus, our models show some signs of in-context learning, as can be seen in \cref{tab:icl_sst2_varying}. When using the causal attention mask, we see that while the models trained with a single objective underperform the baseline, the hybrid models all perform above the majority baseline (from +0.5\% to +5.7\%); with the best results being achieved by the 3:1 model (with the 1:3 and 7:1 close second and third respectively). This indicates that our models are capable of doing in-context learning when trained with both objectives. When run fully bidirectionally, the trend is similar but with lower absolute performance. 




\section{Related work}
\label{sec:related}

\paragraph{Baby language models} This paper describes a submission to the second iteration of the BabyLM challenge
\citep{warstadt-etal-2023-findings}. Our submission is heavily inspired by the last-year's winner, ELC-BERT \citep{georges-gabriel-charpentier-samuel-2023-layers}, and by its inspiration, LTG-BERT \citep{samuel-etal-2023-trained}. Our modifications to these approaches are described in \Cref{sec:hybrid} and \Cref{sec:modification}.

\paragraph{Hybrid masked-causal models} Our work is not the first to attempt to merge bidirectional masked language modeling with generative causal modeling: T5 \citep{10.5555/3455716.3455856}, BART \citep{lewis-etal-2020-bart} and GLM \citep{du-etal-2022-glm} proposed autoregressive fill-in-the-blank training objectives, CM3 is based on a causal-mask objective \citep{aghajanyan2022cm3}, prefix language models use a partially-bidirectional causal modeling \citep{NEURIPS2019_c20bb2d9, 10.5555/3455716.3455856}, and UL2 further improves the T5 encoder-decoder with more training objectives \citep{tay2023ul}. Our approach differs by its simplicity -- not requiring any architectural changes nor novel training objectives -- it just combines a standard causal language model with a (shifted) masked language model; the resulting hybrid can then be used as any GPT-like or BERT-like model out-of-the-box.

\paragraph{Masked next-token prediction} To our best knowledge, this training objective was first proposed by \newcite{DBLP:journals/corr/abs-2311-07468} with some additional modifications, and then simplified in LLM2Vec by \newcite{behnamghader2024llmvec}, where it was used to finetune purely causal language models so that they can function as bidirectional text embedders. Since we use the latter formulation in this paper, we refer to this training objective as `masked next-token prediction'.

\section{Conclusion}
\label{sec:conclusion}

We introduced GPT-BERT, a novel approach that unifies masked and causal language modeling objectives within a single transformer architecture. Through extensive experiments on the BabyLM Challenge 2024, we demonstrated that this hybrid approach offers several key advantages over single-objective models:
\begin{enumerate}[itemsep=0.15em]
    \item \textit{Improved performance:} The hybrid pretraining leads to better results across multiple benchmarks, outperforming both pure MLM and pure CLM approaches.
    \item \textit{Architectural flexibility:} Without any structural modifications, our model can operate in masked, causal, or prefix modes. This flexibility enables GPT-BERT to handle a diverse range of tasks using the most appropriate inference strategy for each situation.
    \item \textit{Unexpected capabilities:} Despite being trained on limited data and having a relatively small parameter count, our models exhibit signs of in-context learning -- a capability typically associated with much larger models.
    \item \textit{Training efficiency:} The hybrid approach achieves these improvements without requiring additional parameters or increased training time compared to single-objective models.
\end{enumerate}

\noindent
Our results suggest that the traditional dichotomy between MLM and CLM architectures may be unnecessary, and that future work might benefit from exploring more unified approaches to language model pretraining.


\section*{Limitations}

While the results presented in this paper are promising and suggest improvements across many tasks when using GPT-BERT, all tested models are relatively small and trained on very small datasets. There is a possibility that these results do not scale and do not work outside of the BabyLM constraints.

\section*{Acknowledgments}

This work is fully funded by the University of Oslo. The computations were performed on resources provided through Sigma2 – the national research infrastructure provider for high-performance computing and large-scale data storage in Norway. We acknowledge Norway and Sigma2 for awarding this project access to the LUMI supercomputer, owned by the EuroHPC Joint Undertaking, hosted by CSC (Finland) and the LUMI consortium through project 5000144.

\bibliography{custom, anthology}

\begin{thebibliography}{52}
\providecommand{\natexlab}[1]{#1}

\bibitem[{Aghajanyan et~al.(2022)Aghajanyan, Huang, Ross, Karpukhin, Xu, Goyal, Okhonko, Joshi, Ghosh, Lewis, and Zettlemoyer}]{aghajanyan2022cm3}
Armen Aghajanyan, Bernie Huang, Candace Ross, Vladimir Karpukhin, Hu~Xu, Naman Goyal, Dmytro Okhonko, Mandar Joshi, Gargi Ghosh, Mike Lewis, and Luke Zettlemoyer. 2022.
\newblock \href {https://arxiv.org/abs/2201.07520} {{CM3}: A causal masked multimodal model of the internet}.
\newblock \emph{Preprint}, arXiv:2201.07520.

\bibitem[{Ankner et~al.(2024)Ankner, Saphra, Blalock, Frankle, and Leavitt}]{ankner-etal-2024-dynamic}
Zachary Ankner, Naomi Saphra, Davis Blalock, Jonathan Frankle, and Matthew Leavitt. 2024.
\newblock \href {https://aclanthology.org/2024.eacl-short.42} {Dynamic masking rate schedules for {MLM} pretraining}.
\newblock In \emph{Proceedings of the 18th Conference of the European Chapter of the Association for Computational Linguistics (Volume 2: Short Papers)}, pages 477--487, St. Julian{'}s, Malta. Association for Computational Linguistics.

\bibitem[{Bar-Haim et~al.(2006)Bar-Haim, Dagan, Dolan, Ferro, and Giampiccolo}]{rte2}
Roy Bar-Haim, Ido Dagan, Bill Dolan, Lisa Ferro, and Danilo Giampiccolo. 2006.
\newblock The second pascal recognising textual entailment challenge.
\newblock \emph{Proceedings of the Second PASCAL Challenges Workshop on Recognising Textual Entailment}.

\bibitem[{BehnamGhader et~al.(2024)BehnamGhader, Adlakha, Mosbach, Bahdanau, Chapados, and Reddy}]{behnamghader2024llmvec}
Parishad BehnamGhader, Vaibhav Adlakha, Marius Mosbach, Dzmitry Bahdanau, Nicolas Chapados, and Siva Reddy. 2024.
\newblock \href {https://openreview.net/forum?id=IW1PR7vEBf} {{LLM}2vec: Large language models are secretly powerful text encoders}.
\newblock In \emph{First Conference on Language Modeling}.

\bibitem[{Belinkov(2022)}]{10.1162/coli_a_00422}
Yonatan Belinkov. 2022.
\newblock \href {https://doi.org/10.1162/coli_a_00422} {{Probing Classifiers: Promises, Shortcomings, and Advances}}.
\newblock \emph{Computational Linguistics}, 48(1):207--219.

\bibitem[{Ben~Allal et~al.(2024)Ben~Allal, Lozhkov, Penedo, Wolf, and von Werra}]{benallal2024cosmopedia}
Loubna Ben~Allal, Anton Lozhkov, Guilherme Penedo, Thomas Wolf, and Leandro von Werra. 2024.
\newblock \href {https://huggingface.co/datasets/HuggingFaceTB/cosmopedia} {Cosmopedia}.

\bibitem[{Bentivogli et~al.(2009)Bentivogli, Dagan, Dang, Giampiccolo, and Magnini}]{Bentivogli09thefifth}
Luisa Bentivogli, Ido Dagan, Hoa~Trang Dang, Danilo Giampiccolo, and Bernardo Magnini. 2009.
\newblock The fifth pascal recognizing textual entailment challenge.
\newblock In \emph{In Proc Text Analysis Conference (TAC’09}.

\bibitem[{Brown et~al.(2020)Brown, Mann, Ryder, Subbiah, Kaplan, Dhariwal, Neelakantan, Shyam, Sastry, Askell, Agarwal, Herbert-Voss, Krueger, Henighan, Child, Ramesh, Ziegler, Wu, Winter, Hesse, Chen, Sigler, Litwin, Gray, Chess, Clark, Berner, McCandlish, Radford, Sutskever, and Amodei}]{NEURIPS2020_1457c0d6}
Tom Brown, Benjamin Mann, Nick Ryder, Melanie Subbiah, Jared~D Kaplan, Prafulla Dhariwal, Arvind Neelakantan, Pranav Shyam, Girish Sastry, Amanda Askell, Sandhini Agarwal, Ariel Herbert-Voss, Gretchen Krueger, Tom Henighan, Rewon Child, Aditya Ramesh, Daniel Ziegler, Jeffrey Wu, Clemens Winter, Chris Hesse, Mark Chen, Eric Sigler, Mateusz Litwin, Scott Gray, Benjamin Chess, Jack Clark, Christopher Berner, Sam McCandlish, Alec Radford, Ilya Sutskever, and Dario Amodei. 2020.
\newblock \href {https://proceedings.neurips.cc/paper_files/paper/2020/file/1457c0d6bfcb4967418bfb8ac142f64a-Paper.pdf} {Language models are few-shot learners}.
\newblock In \emph{Advances in Neural Information Processing Systems}, volume~33, pages 1877--1901. Curran Associates, Inc.

\bibitem[{Choshen et~al.(2024)Choshen, Cotterell, Hu, Linzen, Mueller, Ross, Warstadt, Wilcox, Williams, and Zhuang}]{babylm-2024}
Leshem Choshen, Ryan Cotterell, Michael~Y. Hu, Tal Linzen, Aaron Mueller, Candace Ross, Alex Warstadt, Ethan Wilcox, Adina Williams, and Chengxu Zhuang. 2024.
\newblock \href {https://arxiv.org/abs/2404.06214} {[call for papers] the 2nd {BabyLM} {C}hallenge: Sample-efficient pretraining on a developmentally plausible corpus}.
\newblock \emph{Computing Research Repository}, arXiv:2404.06214.

\bibitem[{Clark et~al.(2019)Clark, Lee, Chang, Kwiatkowski, Collins, and Toutanova}]{clark-etal-2019-boolq}
Christopher Clark, Kenton Lee, Ming-Wei Chang, Tom Kwiatkowski, Michael Collins, and Kristina Toutanova. 2019.
\newblock \href {https://doi.org/10.18653/v1/N19-1300} {{B}ool{Q}: Exploring the surprising difficulty of natural yes/no questions}.
\newblock In \emph{Proceedings of the 2019 Conference of the North {A}merican Chapter of the Association for Computational Linguistics: Human Language Technologies, Volume 1 (Long and Short Papers)}, pages 2924--2936, Minneapolis, Minnesota. Association for Computational Linguistics.

\bibitem[{Dagan et~al.(2006)Dagan, Glickman, and Magnini}]{10.1007/11736790_9}
Ido Dagan, Oren Glickman, and Bernardo Magnini. 2006.
\newblock The pascal recognising textual entailment challenge.
\newblock In \emph{Machine Learning Challenges. Evaluating Predictive Uncertainty, Visual Object Classification, and Recognising Tectual Entailment}, pages 177--190, Berlin, Heidelberg. Springer Berlin Heidelberg.

\bibitem[{DeepSeek-AI(2024)}]{deepseekv2}
DeepSeek-AI. 2024.
\newblock \href {https://arxiv.org/abs/2405.04434} {Deepseek-v2: A strong, economical, and efficient mixture-of-experts language model}.
\newblock \emph{Preprint}, arXiv:2405.04434.

\bibitem[{Devlin et~al.(2019)Devlin, Chang, Lee, and Toutanova}]{devlin-etal-2019-bert}
Jacob Devlin, Ming-Wei Chang, Kenton Lee, and Kristina Toutanova. 2019.
\newblock \href {https://doi.org/10.18653/v1/N19-1423} {{BERT}: Pre-training of deep bidirectional transformers for language understanding}.
\newblock In \emph{Proceedings of the 2019 Conference of the North {A}merican Chapter of the Association for Computational Linguistics: Human Language Technologies, Volume 1 (Long and Short Papers)}, pages 4171--4186, Minneapolis, Minnesota. Association for Computational Linguistics.

\bibitem[{Dolan and Brockett(2005)}]{dolan-brockett-2005-automatically}
William~B. Dolan and Chris Brockett. 2005.
\newblock \href {https://aclanthology.org/I05-5002} {Automatically constructing a corpus of sentential paraphrases}.
\newblock In \emph{Proceedings of the Third International Workshop on Paraphrasing ({IWP}2005)}.

\bibitem[{Dong et~al.(2019)Dong, Yang, Wang, Wei, Liu, Wang, Gao, Zhou, and Hon}]{NEURIPS2019_c20bb2d9}
Li~Dong, Nan Yang, Wenhui Wang, Furu Wei, Xiaodong Liu, Yu~Wang, Jianfeng Gao, Ming Zhou, and Hsiao-Wuen Hon. 2019.
\newblock \href {https://proceedings.neurips.cc/paper_files/paper/2019/file/c20bb2d9a50d5ac1f713f8b34d9aac5a-Paper.pdf} {Unified language model pre-training for natural language understanding and generation}.
\newblock In \emph{Advances in Neural Information Processing Systems}, volume~32. Curran Associates, Inc.

\bibitem[{Du et~al.(2022)Du, Qian, Liu, Ding, Qiu, Yang, and Tang}]{du-etal-2022-glm}
Zhengxiao Du, Yujie Qian, Xiao Liu, Ming Ding, Jiezhong Qiu, Zhilin Yang, and Jie Tang. 2022.
\newblock \href {https://doi.org/10.18653/v1/2022.acl-long.26} {{GLM}: General language model pretraining with autoregressive blank infilling}.
\newblock In \emph{Proceedings of the 60th Annual Meeting of the Association for Computational Linguistics (Volume 1: Long Papers)}, pages 320--335, Dublin, Ireland. Association for Computational Linguistics.

\bibitem[{Ewer et~al.(2024)Ewer, Chae, Zeng, Kim, and Lee}]{ewer2024entpencoderonlytokenprediction}
Ethan Ewer, Daewon Chae, Thomas Zeng, Jinkyu Kim, and Kangwook Lee. 2024.
\newblock \href {https://arxiv.org/abs/2410.01600} {Entp: Encoder-only next token prediction}.
\newblock \emph{Preprint}, arXiv:2410.01600.

\bibitem[{Gage(1994)}]{Gage1994ANA}
Philip Gage. 1994.
\newblock \href {https://api.semanticscholar.org/CorpusID:59804030} {A new algorithm for data compression}.
\newblock \emph{The C Users Journal archive}, 12:23--38.

\bibitem[{Gao et~al.(2023)Gao, Tow, Abbasi, Biderman, Black, DiPofi, Foster, Golding, Hsu, Le~Noac'h, Li, McDonell, Muennighoff, Ociepa, Phang, Reynolds, Schoelkopf, Skowron, Sutawika, Tang, Thite, Wang, Wang, and Zou}]{eval-harness}
Leo Gao, Jonathan Tow, Baber Abbasi, Stella Biderman, Sid Black, Anthony DiPofi, Charles Foster, Laurence Golding, Jeffrey Hsu, Alain Le~Noac'h, Haonan Li, Kyle McDonell, Niklas Muennighoff, Chris Ociepa, Jason Phang, Laria Reynolds, Hailey Schoelkopf, Aviya Skowron, Lintang Sutawika, Eric Tang, Anish Thite, Ben Wang, Kevin Wang, and Andy Zou. 2023.
\newblock \href {https://doi.org/10.5281/zenodo.10256836} {A framework for few-shot language model evaluation}.

\bibitem[{Georges Gabriel~Charpentier and Samuel(2023)}]{georges-gabriel-charpentier-samuel-2023-layers}
Lucas Georges Gabriel~Charpentier and David Samuel. 2023.
\newblock \href {https://doi.org/10.18653/v1/2023.conll-babylm.20} {Not all layers are equally as important: Every layer counts {BERT}}.
\newblock In \emph{Proceedings of the BabyLM Challenge at the 27th Conference on Computational Natural Language Learning}, pages 238--252, Singapore. Association for Computational Linguistics.

\bibitem[{Giampiccolo et~al.(2007)Giampiccolo, Magnini, Dagan, and Dolan}]{giampiccolo-etal-2007-third}
Danilo Giampiccolo, Bernardo Magnini, Ido Dagan, and Bill Dolan. 2007.
\newblock \href {https://aclanthology.org/W07-1401} {The third {PASCAL} recognizing textual entailment challenge}.
\newblock In \emph{Proceedings of the {ACL}-{PASCAL} Workshop on Textual Entailment and Paraphrasing}, pages 1--9, Prague. Association for Computational Linguistics.

\bibitem[{Guo et~al.(2017)Guo, Pleiss, Sun, and Weinberger}]{10.5555/3305381.3305518}
Chuan Guo, Geoff Pleiss, Yu~Sun, and Kilian~Q. Weinberger. 2017.
\newblock On calibration of modern neural networks.
\newblock In \emph{Proceedings of the 34th International Conference on Machine Learning - Volume 70}, ICML'17, page 1321–1330. JMLR.org.

\bibitem[{Ivanova et~al.(2024)Ivanova, Sathe, Lipkin, Kumar, Radkani, Clark, Kauf, Hu, Pramod, Grand, Paulun, Ryskina, Akyürek, Wilcox, Rashid, Choshen, Levy, Fedorenko, Tenenbaum, and Andreas}]{ivanova2024elementsworldknowledgeewok}
Anna~A. Ivanova, Aalok Sathe, Benjamin Lipkin, Unnathi Kumar, Setayesh Radkani, Thomas~H. Clark, Carina Kauf, Jennifer Hu, R.~T. Pramod, Gabriel Grand, Vivian Paulun, Maria Ryskina, Ekin Akyürek, Ethan Wilcox, Nafisa Rashid, Leshem Choshen, Roger Levy, Evelina Fedorenko, Joshua Tenenbaum, and Jacob Andreas. 2024.
\newblock \href {https://arxiv.org/abs/2405.09605} {Elements of world knowledge (ewok): A cognition-inspired framework for evaluating basic world knowledge in language models}.
\newblock \emph{Preprint}, arXiv:2405.09605.

\bibitem[{Jumper et~al.(2021)Jumper, Evans, Pritzel, Green, Figurnov, Ronneberger, Tunyasuvunakool, Bates, {\v{Z}}{\'\i}dek, Potapenko, Bridgland, Meyer, Kohl, Ballard, Cowie, Romera-Paredes, Nikolov, Jain, Adler, Back, Petersen, Reiman, Clancy, Zielinski, Steinegger, Pacholska, Berghammer, Bodenstein, Silver, Vinyals, Senior, Kavukcuoglu, Kohli, and Hassabis}]{AlphaFold2021}
John Jumper, Richard Evans, Alexander Pritzel, Tim Green, Michael Figurnov, Olaf Ronneberger, Kathryn Tunyasuvunakool, Russ Bates, Augustin {\v{Z}}{\'\i}dek, Anna Potapenko, Alex Bridgland, Clemens Meyer, Simon A~A Kohl, Andrew~J Ballard, Andrew Cowie, Bernardino Romera-Paredes, Stanislav Nikolov, Rishub Jain, Jonas Adler, Trevor Back, Stig Petersen, David Reiman, Ellen Clancy, Michal Zielinski, Martin Steinegger, Michalina Pacholska, Tamas Berghammer, Sebastian Bodenstein, David Silver, Oriol Vinyals, Andrew~W Senior, Koray Kavukcuoglu, Pushmeet Kohli, and Demis Hassabis. 2021.
\newblock \href {https://doi.org/10.1038/s41586-021-03819-2} {Highly accurate protein structure prediction with {AlphaFold}}.
\newblock \emph{Nature}, 596(7873):583--589.

\bibitem[{Khashabi et~al.(2018)Khashabi, Chaturvedi, Roth, Upadhyay, and Roth}]{khashabi-etal-2018-looking}
Daniel Khashabi, Snigdha Chaturvedi, Michael Roth, Shyam Upadhyay, and Dan Roth. 2018.
\newblock \href {https://doi.org/10.18653/v1/N18-1023} {Looking beyond the surface: A challenge set for reading comprehension over multiple sentences}.
\newblock In \emph{Proceedings of the 2018 Conference of the North {A}merican Chapter of the Association for Computational Linguistics: Human Language Technologies, Volume 1 (Long Papers)}, pages 252--262, New Orleans, Louisiana. Association for Computational Linguistics.

\bibitem[{Levesque et~al.(2012)Levesque, Davis, and Morgenstern}]{10.5555/3031843.3031909}
Hector~J. Levesque, Ernest Davis, and Leora Morgenstern. 2012.
\newblock The winograd schema challenge.
\newblock In \emph{Proceedings of the Thirteenth International Conference on Principles of Knowledge Representation and Reasoning}, KR'12, page 552–561. AAAI Press.

\bibitem[{Lewis et~al.(2020)Lewis, Liu, Goyal, Ghazvininejad, Mohamed, Levy, Stoyanov, and Zettlemoyer}]{lewis-etal-2020-bart}
Mike Lewis, Yinhan Liu, Naman Goyal, Marjan Ghazvininejad, Abdelrahman Mohamed, Omer Levy, Veselin Stoyanov, and Luke Zettlemoyer. 2020.
\newblock \href {https://doi.org/10.18653/v1/2020.acl-main.703} {{BART}: Denoising sequence-to-sequence pre-training for natural language generation, translation, and comprehension}.
\newblock In \emph{Proceedings of the 58th Annual Meeting of the Association for Computational Linguistics}, pages 7871--7880, Online. Association for Computational Linguistics.

\bibitem[{Lozhkov et~al.(2024)Lozhkov, Ben~Allal, von Werra, and Wolf}]{lozhkov2024fineweb-edu}
Anton Lozhkov, Loubna Ben~Allal, Leandro von Werra, and Thomas Wolf. 2024.
\newblock \href {https://doi.org/10.57967/hf/2497} {Fineweb-edu}.

\bibitem[{Lv et~al.(2023)Lv, Zhang, Xie, Tu, Chen, Wen, and Yan}]{DBLP:journals/corr/abs-2311-07468}
Ang Lv, Kaiyi Zhang, Shufang Xie, Quan Tu, Yuhan Chen, Ji-Rong Wen, and Rui Yan. 2023.
\newblock \href {https://doi.org/10.48550/arXiv.2311.07468} {Are we falling in a middle-intelligence trap? an analysis and mitigation of the reversal curse}.
\newblock \emph{CoRR}, abs/2311.07468.

\bibitem[{Matthews(1975)}]{MATTHEWS1975442}
B.W. Matthews. 1975.
\newblock \href {https://doi.org/10.1016/0005-2795(75)90109-9} {Comparison of the predicted and observed secondary structure of t4 phage lysozyme}.
\newblock \emph{Biochimica et Biophysica Acta (BBA) - Protein Structure}, 405(2):442--451.

\bibitem[{Pagliardini et~al.(2024)Pagliardini, Mohtashami, Fleuret, and Jaggi}]{pagliardini2024denseformer}
Matteo Pagliardini, Amirkeivan Mohtashami, Francois Fleuret, and Martin Jaggi. 2024.
\newblock Denseformer: Enhancing information flow in transformers via depth weighted averaging.
\newblock \emph{arXiv preprint arXiv:2402.02622}.

\bibitem[{Paperno et~al.(2016)Paperno, Kruszewski, Lazaridou, Pham, Bernardi, Pezzelle, Baroni, Boleda, and Fern{\'a}ndez}]{paperno-etal-2016-lambada}
Denis Paperno, Germ{\'a}n Kruszewski, Angeliki Lazaridou, Ngoc~Quan Pham, Raffaella Bernardi, Sandro Pezzelle, Marco Baroni, Gemma Boleda, and Raquel Fern{\'a}ndez. 2016.
\newblock \href {https://doi.org/10.18653/v1/P16-1144} {The {LAMBADA} dataset: Word prediction requiring a broad discourse context}.
\newblock In \emph{Proceedings of the 54th Annual Meeting of the Association for Computational Linguistics (Volume 1: Long Papers)}, pages 1525--1534, Berlin, Germany. Association for Computational Linguistics.

\bibitem[{Radford et~al.(2018)Radford, Narasimhan, Salimans, and Sutskever}]{Radford2018ImprovingLU}
Alec Radford, Karthik Narasimhan, Tim Salimans, and Ilya Sutskever. 2018.
\newblock \href {https://cdn.openai.com/research-covers/language-unsupervised/language_understanding_paper.pdf} {Improving language understanding by generative pre-training}.

\bibitem[{Radford et~al.(2019)Radford, Wu, Child, Luan, Amodei, Sutskever et~al.}]{radford2019language}
Alec Radford, Jeffrey Wu, Rewon Child, David Luan, Dario Amodei, Ilya Sutskever, et~al. 2019.
\newblock Language models are unsupervised multitask learners.
\newblock \emph{OpenAI blog}, 1(8):9.

\bibitem[{Rae et~al.(2022)Rae, Borgeaud, Cai, Millican, Hoffmann, Song, Aslanides, Henderson, Ring, Young, Rutherford, Hennigan, Menick, Cassirer, Powell, van~den Driessche, Hendricks, Rauh, Huang, Glaese, Welbl, Dathathri, Huang, Uesato, Mellor, Higgins, Creswell, McAleese, Wu, Elsen, Jayakumar, Buchatskaya, Budden, Sutherland, Simonyan, Paganini, Sifre, Martens, Li, Kuncoro, Nematzadeh, Gribovskaya, Donato, Lazaridou, Mensch, Lespiau, Tsimpoukelli, Grigorev, Fritz, Sottiaux, Pajarskas, Pohlen, Gong, Toyama, de~Masson~d'Autume, Li, Terzi, Mikulik, Babuschkin, Clark, de~Las~Casas, Guy, Jones, Bradbury, Johnson, Hechtman, Weidinger, Gabriel, Isaac, Lockhart, Osindero, Rimell, Dyer, Vinyals, Ayoub, Stanway, Bennett, Hassabis, Kavukcuoglu, and Irving}]{rae2022scalinglanguagemodelsmethods}
Jack~W. Rae, Sebastian Borgeaud, Trevor Cai, Katie Millican, Jordan Hoffmann, Francis Song, John Aslanides, Sarah Henderson, Roman Ring, Susannah Young, Eliza Rutherford, Tom Hennigan, Jacob Menick, Albin Cassirer, Richard Powell, George van~den Driessche, Lisa~Anne Hendricks, Maribeth Rauh, Po-Sen Huang, Amelia Glaese, Johannes Welbl, Sumanth Dathathri, Saffron Huang, Jonathan Uesato, John Mellor, Irina Higgins, Antonia Creswell, Nat McAleese, Amy Wu, Erich Elsen, Siddhant Jayakumar, Elena Buchatskaya, David Budden, Esme Sutherland, Karen Simonyan, Michela Paganini, Laurent Sifre, Lena Martens, Xiang~Lorraine Li, Adhiguna Kuncoro, Aida Nematzadeh, Elena Gribovskaya, Domenic Donato, Angeliki Lazaridou, Arthur Mensch, Jean-Baptiste Lespiau, Maria Tsimpoukelli, Nikolai Grigorev, Doug Fritz, Thibault Sottiaux, Mantas Pajarskas, Toby Pohlen, Zhitao Gong, Daniel Toyama, Cyprien de~Masson~d'Autume, Yujia Li, Tayfun Terzi, Vladimir Mikulik, Igor Babuschkin, Aidan Clark, Diego de~Las~Casas, Aurelia Guy, Chris Jones,
  James Bradbury, Matthew Johnson, Blake Hechtman, Laura Weidinger, Iason Gabriel, William Isaac, Ed~Lockhart, Simon Osindero, Laura Rimell, Chris Dyer, Oriol Vinyals, Kareem Ayoub, Jeff Stanway, Lorrayne Bennett, Demis Hassabis, Koray Kavukcuoglu, and Geoffrey Irving. 2022.
\newblock \href {https://arxiv.org/abs/2112.11446} {Scaling language models: Methods, analysis \& insights from training gopher}.
\newblock \emph{Preprint}, arXiv:2112.11446.

\bibitem[{Raffel et~al.(2020)Raffel, Shazeer, Roberts, Lee, Narang, Matena, Zhou, Li, and Liu}]{10.5555/3455716.3455856}
Colin Raffel, Noam Shazeer, Adam Roberts, Katherine Lee, Sharan Narang, Michael Matena, Yanqi Zhou, Wei Li, and Peter~J. Liu. 2020.
\newblock \href {https://www.jmlr.org/papers/volume21/20-074/20-074.pdf} {Exploring the limits of transfer learning with a unified text-to-text transformer}.
\newblock \emph{J. Mach. Learn. Res.}, 21(1).

\bibitem[{Rajpurkar et~al.(2016)Rajpurkar, Zhang, Lopyrev, and Liang}]{rajpurkar-etal-2016-squad}
Pranav Rajpurkar, Jian Zhang, Konstantin Lopyrev, and Percy Liang. 2016.
\newblock \href {https://doi.org/10.18653/v1/D16-1264} {{SQ}u{AD}: 100,000+ questions for machine comprehension of text}.
\newblock In \emph{Proceedings of the 2016 Conference on Empirical Methods in Natural Language Processing}, pages 2383--2392, Austin, Texas. Association for Computational Linguistics.

\bibitem[{Salazar et~al.(2020)Salazar, Liang, Nguyen, and Kirchhoff}]{salazar-etal-2020-masked}
Julian Salazar, Davis Liang, Toan~Q. Nguyen, and Katrin Kirchhoff. 2020.
\newblock \href {https://doi.org/10.18653/v1/2020.acl-main.240} {Masked language model scoring}.
\newblock In \emph{Proceedings of the 58th Annual Meeting of the Association for Computational Linguistics}, pages 2699--2712, Online. Association for Computational Linguistics.

\bibitem[{Samuel(2024)}]{samuel2024berts}
David Samuel. 2024.
\newblock \href {https://openreview.net/forum?id=BCA9NMZkLS} {{BERT}s are generative in-context learners}.
\newblock In \emph{The Thirty-eighth Annual Conference on Neural Information Processing Systems}.

\bibitem[{Samuel et~al.(2023)Samuel, Kutuzov, {\O}vrelid, and Velldal}]{samuel-etal-2023-trained}
David Samuel, Andrey Kutuzov, Lilja {\O}vrelid, and Erik Velldal. 2023.
\newblock \href {https://doi.org/10.18653/v1/2023.findings-eacl.146} {Trained on 100 million words and still in shape: {BERT} meets {B}ritish {N}ational {C}orpus}.
\newblock In \emph{Findings of the Association for Computational Linguistics: EACL 2023}, pages 1954--1974, Dubrovnik, Croatia. Association for Computational Linguistics.

\bibitem[{Shazeer(2020)}]{shazeer2020gluvariantsimprovetransformer}
Noam Shazeer. 2020.
\newblock \href {https://arxiv.org/abs/2002.05202} {Glu variants improve transformer}.
\newblock \emph{Preprint}, arXiv:2002.05202.

\bibitem[{Socher et~al.(2013)Socher, Perelygin, Wu, Chuang, Manning, Ng, and Potts}]{socher-etal-2013-recursive}
Richard Socher, Alex Perelygin, Jean Wu, Jason Chuang, Christopher~D. Manning, Andrew Ng, and Christopher Potts. 2013.
\newblock \href {https://aclanthology.org/D13-1170} {Recursive deep models for semantic compositionality over a sentiment treebank}.
\newblock In \emph{Proceedings of the 2013 Conference on Empirical Methods in Natural Language Processing}, pages 1631--1642, Seattle, Washington, USA. Association for Computational Linguistics.

\bibitem[{Tay et~al.(2023)Tay, Dehghani, Tran, Garcia, Wei, Wang, Chung, Bahri, Schuster, Zheng, Zhou, Houlsby, and Metzler}]{tay2023ul}
Yi~Tay, Mostafa Dehghani, Vinh~Q. Tran, Xavier Garcia, Jason Wei, Xuezhi Wang, Hyung~Won Chung, Dara Bahri, Tal Schuster, Steven Zheng, Denny Zhou, Neil Houlsby, and Donald Metzler. 2023.
\newblock \href {https://openreview.net/forum?id=6ruVLB727MC} {{UL}2: Unifying language learning paradigms}.
\newblock In \emph{The Eleventh International Conference on Learning Representations}.

\bibitem[{Timiryasov and Tastet(2023)}]{timiryasov-tastet-2023-baby}
Inar Timiryasov and Jean-Loup Tastet. 2023.
\newblock \href {https://doi.org/10.18653/v1/2023.conll-babylm.24} {Baby llama: knowledge distillation from an ensemble of teachers trained on a small dataset with no performance penalty}.
\newblock In \emph{Proceedings of the BabyLM Challenge at the 27th Conference on Computational Natural Language Learning}, pages 279--289, Singapore. Association for Computational Linguistics.

\bibitem[{Wang and Cho(2019)}]{wang-cho-2019-bert}
Alex Wang and Kyunghyun Cho. 2019.
\newblock \href {https://doi.org/10.18653/v1/W19-2304} {{BERT} has a mouth, and it must speak: {BERT} as a {M}arkov random field language model}.
\newblock In \emph{Proceedings of the Workshop on Methods for Optimizing and Evaluating Neural Language Generation}, pages 30--36, Minneapolis, Minnesota. Association for Computational Linguistics.

\bibitem[{Wang et~al.(2019)Wang, Pruksachatkun, Nangia, Singh, Michael, Hill, Levy, and Bowman}]{NEURIPS2019_4496bf24}
Alex Wang, Yada Pruksachatkun, Nikita Nangia, Amanpreet Singh, Julian Michael, Felix Hill, Omer Levy, and Samuel Bowman. 2019.
\newblock \href {https://proceedings.neurips.cc/paper/2019/file/4496bf24afe7fab6f046bf4923da8de6-Paper.pdf} {Superglue: A stickier benchmark for general-purpose language understanding systems}.
\newblock In \emph{Advances in Neural Information Processing Systems}, volume~32. Curran Associates, Inc.

\bibitem[{Wang et~al.(2018)Wang, Singh, Michael, Hill, Levy, and Bowman}]{wang-etal-2018-glue}
Alex Wang, Amanpreet Singh, Julian Michael, Felix Hill, Omer Levy, and Samuel Bowman. 2018.
\newblock \href {https://doi.org/10.18653/v1/W18-5446} {{GLUE}: A multi-task benchmark and analysis platform for natural language understanding}.
\newblock In \emph{Proceedings of the 2018 {EMNLP} Workshop {B}lackbox{NLP}: Analyzing and Interpreting Neural Networks for {NLP}}, pages 353--355, Brussels, Belgium. Association for Computational Linguistics.

\bibitem[{Warstadt et~al.(2023)Warstadt, Mueller, Choshen, Wilcox, Zhuang, Ciro, Mosquera, Paranjabe, Williams, Linzen, and Cotterell}]{warstadt-etal-2023-findings}
Alex Warstadt, Aaron Mueller, Leshem Choshen, Ethan Wilcox, Chengxu Zhuang, Juan Ciro, Rafael Mosquera, Bhargavi Paranjabe, Adina Williams, Tal Linzen, and Ryan Cotterell. 2023.
\newblock \href {https://doi.org/10.18653/v1/2023.conll-babylm.1} {Findings of the {B}aby{LM} challenge: Sample-efficient pretraining on developmentally plausible corpora}.
\newblock In \emph{Proceedings of the BabyLM Challenge at the 27th Conference on Computational Natural Language Learning}, pages 1--34, Singapore. Association for Computational Linguistics.

\bibitem[{Warstadt et~al.(2020)Warstadt, Parrish, Liu, Mohananey, Peng, Wang, and Bowman}]{warstadt-etal-2020-blimp-benchmark}
Alex Warstadt, Alicia Parrish, Haokun Liu, Anhad Mohananey, Wei Peng, Sheng-Fu Wang, and Samuel~R. Bowman. 2020.
\newblock \href {https://doi.org/10.1162/tacl_a_00321} {{BL}i{MP}: The benchmark of linguistic minimal pairs for {E}nglish}.
\newblock \emph{Transactions of the Association for Computational Linguistics}, 8:377--392.

\bibitem[{Warstadt et~al.(2019)Warstadt, Singh, and Bowman}]{warstadt-etal-2019-neural}
Alex Warstadt, Amanpreet Singh, and Samuel~R. Bowman. 2019.
\newblock \href {https://doi.org/10.1162/tacl_a_00290} {Neural network acceptability judgments}.
\newblock \emph{Transactions of the Association for Computational Linguistics}, 7:625--641.

\bibitem[{Wei et~al.(2022)Wei, Tay, Bommasani, Raffel, Zoph, Borgeaud, Yogatama, Bosma, Zhou, Metzler, Chi, Hashimoto, Vinyals, Liang, Dean, and Fedus}]{wei2022emergent}
Jason Wei, Yi~Tay, Rishi Bommasani, Colin Raffel, Barret Zoph, Sebastian Borgeaud, Dani Yogatama, Maarten Bosma, Denny Zhou, Donald Metzler, Ed~H. Chi, Tatsunori Hashimoto, Oriol Vinyals, Percy Liang, Jeff Dean, and William Fedus. 2022.
\newblock \href {https://openreview.net/forum?id=yzkSU5zdwD} {Emergent abilities of large language models}.
\newblock \emph{Transactions on Machine Learning Research}.
\newblock Survey Certification.

\bibitem[{Williams et~al.(2018)Williams, Nangia, and Bowman}]{williams-etal-2018-broad}
Adina Williams, Nikita Nangia, and Samuel Bowman. 2018.
\newblock \href {https://doi.org/10.18653/v1/N18-1101} {A broad-coverage challenge corpus for sentence understanding through inference}.
\newblock In \emph{Proceedings of the 2018 Conference of the North {A}merican Chapter of the Association for Computational Linguistics: Human Language Technologies, Volume 1 (Long Papers)}, pages 1112--1122, New Orleans, Louisiana. Association for Computational Linguistics.

\end{thebibliography}

\clearpage
\onecolumn
\appendix

\section{Pre-training details} \label{app:training}

\begin{table*}[ht!]
\centering
\small
\begin{tabular}{@{}lccc@{}}
\toprule
\textbf{Hyperparameter} & \textbf{\textsc{strict} (100M)} & \textbf{\textsc{strict-small} (10M)} \\ \midrule
Number of parameters    & 119M & 30M \\
Number of layers$^\dagger$        & 12 & 12       \\
Hidden size             & 768 & 384       \\
FF intermediate size    & 2\,560 & 1\,280 \\
Vocabulary size         & 16\,384 & 8\,192        \\
Attention heads         & 12 & 6        \\
Hidden dropout                 & 0.1   & 0.1      \\
Attention dropout       & 0.1   & 0.1        \\
Training steps          & 15\,625 & 7\,812   \\
Batch size              & 1M $\to$ 4M (tokens)& 1M $\to$ 4M (tokens)     \\
Initial Sequence length         & 128 & 128      \\
Final Sequence length         & 512 & 512      \\
Warmup ratio            & 1.6\% & 1.6\%       \\
Initial learning rate   & 0.01 & 0.0141       \\
Final learning rate     & 0.001 & 0.00141      \\
Learning rate scheduler & cosine & cosine   \\
Weight decay            & 0.1 & 0.1\\
Optimizer               & LAMB & LAMB   \\
LAMB $\epsilon$         & 1e-8 & 1e-8       \\
LAMB $\beta_1$          & 0.9  & 0.9      \\
LAMB $\beta_2$          & 0.98   & 0.98     \\
Gradient clipping       & 2.0  & 2.0        \\ \bottomrule
\end{tabular} %
\caption{\textbf{Pre-training hyperparameters}\hspace{1.5em}We train base-sized models on the \textsc{strict} corpus and small-sized models on the \textsc{strict-small} corpus. $^\dagger$ Here one `layer' refers to one module composed of both the attention and feed-forward submodules; a more standard definition than the one used in \Cref{sec:modification}.}
\label{tab:hyperparams}
\end{table*}

\section{Evaluation details}
\label{app:evaluation}

\paragraph{Hyperparameters} TO find the hyperparameters we do a hyperparameters search on CoLA for the task with small amounts of training data (CoLA, RTE, MRPC, MultiRC) and on SST-2 for tasks with large amounts of training data (QQP, MNLI, QNLI, BoolQ, and SST-2). We do a grid search with values: 
\begin{itemize}
    \item Number of epochs: $\{3, 5, 10\}$
    \item Learning rate: $\{3\times10^{-5}, 5\times10^{-5}, 1\times10^{-4}, 2\times10^{-4}\}$
    \item Batch size: $\{16, 32, 64\}$
\end{itemize}

In addition for WSC given the very low amount of both train and validation data, we expand the search to:
\begin{itemize}
    \item Number of epochs: $\{3, 5, 10, 15, 20, 25, 30, 100\}$
    \item Learning rate: $\{3\times10^{-5}, 5\times10^{-5}, 7\times10^{-5}, 1\times10^{-4}, 2\times10^{-4}, 3\times10^{-4}, 5\times10^{-4}\}$
    \item Batch size: $\{16, 32, 64\}$
    \item Warmup ratio: $\{0.00, 0.06, 0.15\}$
\end{itemize}

The final hyperparameters can be found in \cref{tab:hyperparams_fine}. For MultiRC, we reduce the number of epochs due to the training time.

\begin{table*}[ht!]
\centering
\small
\begin{tabular}{lcccc@{}}
\toprule
\multirow{2}{*}{\textbf{Hyperparameter}} & \textbf{QQP, MNLI, SST-2,} & \multirow{2}{*}{\textbf{CoLA, RTE, MRPC}} & \multirow{2}{*}{\textbf{MultiRC}} & \multirow{2}{*}{\textbf{WSC}} \\
& \textbf{BoolQ, QNLI} \\
\midrule
\multicolumn{1}{@{}l}{\textsc{strict-small}} \\
Number of epochs & 3 & 10 & 3 & 20 \\
Learning rate & $1\times10^{-4}$ & $1\times10^{-4}$ & $1\times10^{-4}$ & $3\times10^{-4}$ \\
Batch size & 16 & 16 & 16 & 32 \\
Warmup ratio & 0.06 & 0.06 & 0.06 & 0.00 \\
Weight decay & 0.01 & 0.01 & 0.01 & 0.01 \\[0.5em]
\multicolumn{1}{@{}l}{\textsc{strict}} \\
Number of epochs & 3 & 10 & 3 & 20 \\
Learning rate & $1\times10^{-4}$ & $1\times10^{-4}$ & $1\times10^{-4}$ & $3\times10^{-4}$ \\
Batch size & 32 & 32 & 32 & 16 \\
Warmup ratio & 0.06 & 0.06 & 0.06 & 0.06 \\
Weight decay & 0.01 & 0.01 & 0.01 & 0.01 \\
 \bottomrule
\end{tabular} %
\caption{\textbf{Fine-tuning hyperparameters}\hspace{1.5em}We use the hyperparameters above to fine-tune our models. We did a hyperparamter search on CoLA and SST-2 to obtain the hyperparameters. For MultiRC, we used less epochs due to the time required to fine-tuned.}
\label{tab:hyperparams_fine}
\end{table*}

\paragraph{(Super)GLUE benchmark.}

General Language Understanding Evaluation benchmarks \citep[GLUE and SuperGLUE;][]{wang-etal-2018-glue, NEURIPS2019_4496bf24} are arguably the most common ways of evaluating the language-understanding and transfer-learning capabilities of language models. The BabyLM challenge uses a subset of 10 (Super)GLUE tasks, detailed in \cref{app:glue}. We employ the standard way of finetuning masked language models on these datasets, as introduced in BERT \citep{devlin-etal-2019-bert}. 

As we use the BabyLM version of GLUE, our results cannot be directly compared with previous literature -- the dataset samples are filtered to not contain out-of-vocabulary words and some of the employed metrics differ from the original recommendations \citep{wang-etal-2018-glue, NEURIPS2019_4496bf24}. We opted to adhere to the BabyLM version to be compatible with other works in this challenge.\footnote{The BabyLM pipeline unfortunately uses identical validation and test sets, which might yield overly optimistic results due to overfitting during hyperparameter optimization.} 

\paragraph{BLiMP.}

When using any finetuning approach, it becomes unclear how to disentangle innate language understanding from knowledge learned during second-stage supervised finetuning \cite{10.1162/coli_a_00422}. In contrast, the Benchmark of Linguistic Minimal Pairs \citep[BLiMP;][]{warstadt-etal-2020-blimp-benchmark} attempts to measure the linguistic knowledge of a language model in a zero-shot manner -- without any additional training. Each pair of sentences in BLiMP differs minimally on the surface level, but only one of the sentences is grammatically valid. We can use the intrinsic ability of language models to assign a probability to every sentence and test how often a language model assigns a higher probability to the correct sentence \citep{wang-cho-2019-bert, salazar-etal-2020-masked}.

As detailed in \cref{app:blimp}, the results on BLiMP greatly depend on temperature scaling \citep{10.5555/3305381.3305518}. Thus, to fairly compare different types of language models, we employ an alternative approach to evaluating BLiMP: we report the accuracies that are achieved with the optimal temperature for every language model; the reasoning is explained in \cref{app:blimp}.

The BabyLM challenge also comes with an additional `BLiMP supplement' held-out set with five additional diagnostic tasks. To comply with the held-out spirit of these tasks, we keep the temperature values calibrated for BLiMP, even though this results in suboptimal performance (\cref{app:blimp}).

\paragraph{EWoK.} Elements of World Knowledge \citep[EWoK;][]{ivanova2024elementsworldknowledgeewok} is another benchmark evaluated in a zero-shot manner. The goal of this benchmark is to evaluate how well a model understand concepts that help model the world. To do this, each example consists of two contexts, and two targets. The first context implicates the first target, and the second context implicates the second target. Both implications are based on world knowledge, such as an object both accelerating and slowing down (context) is oscillating (target).

\paragraph{LAMBADA.} LAnguage Modeling Broadened to Account for Discourse Aspect \citep[LAMBADA;][]{paperno-etal-2016-lambada} is a dataset that verifies how well a language model understand context and whether it is able to recall far-away context. To do this the model must predict an answer token(s) which usually consists of a person, action, fact, or event mentioned earlier in the context.

\newpage

\section{Text Generation with prompts from \citet{radford2019language}}
\label{app:generation}

\begin{table*}[h]
    \centering
    \begin{tabular}{|l|}
        \hline
         \parbox{\textwidth}{%
            \begin{center}
                \textbf{Context}
            \end{center}
            In a shocking finding, scientist discovered a herd of unicorns living in a remote, previously unexplored valley, in the Andes Mountains. Even more surprising to the researchers was the fact that the unicorns spoke perfect English.
        }%
        \\
        \\
         \hline
         \parbox{\textwidth}{%
            \begin{center}
                \textbf{Completion (no repetition penalty)}
            \end{center}
            The unicorns were known for their ability to communicate with each other, and they were known to have a unique way of communicating.
    
            The unicorns were known to have a unique way of communicating with each other, and they were known to have a unique way of communicating with each other.
            
            The unicorns were known to have a unique way of communicating with each other, and they were known to have a unique way of communicating with each other.
            
            The unicorns were known to have a unique way of communicating with each other, and they were known to have a unique way of communicating with each other.
            
            The unicorns were known to have a unique way of communicating with each other, and they were known to have a unique way of communicating with each other.
            
            The unicorns were known to have a unique way of communicating with each other, and they were known to have a unique way of communicating with each other.
            
            The unicorns were known to have a unique way of communicating with each other, and they were known to have a unique way of communicating with each other.
        }%
        \\
        \\
        \hline
        \parbox{\textwidth}{%
            \begin{center}
                \textbf{Completion (1.5 repetition penalty)}
            \end{center}
            The study is published in Nature Communications.
            
            "The unicorn language has been known for centuries as an ancient communication system," said Dr. Mark Rosenberg, lead author of the paper and co-author of the new research. "It's not just a myth; it's also a real one."
            
            A group of scientists from the University of California at Berkeley (UCB) and the University of Colorado at Boulder (UCSD) have found evidence of this phenomenon in some animals. They say they've identified two species of unicorns with similar sounds, but they're different in their own way.
            
            "We are looking at how these unicorns communicate," says Dr. Rosenberg. "They use sound waves to send messages through the air, which then travel down the animal's throat, where it gets picked up by other animals, and back to the brain."
            
            Researchers believe the unicorns' ability to hear can be explained by the presence of certain chemicals called acetylcholinesterase. Acetylcholinesterase helps them detect danger, such as wind or water. It also makes them feel safe when there's a threat.
        }%
        \\
        \\
        \hline
    \end{tabular}
\end{table*}

\begin{table*}[]
    \centering
    \begin{tabular}{|l|}
        \hline
         \parbox{\textwidth}{%
            \begin{center}
                \textbf{Context}
            \end{center}
            Aaron loves mint chocolate cake, but he requires that it be paired with mini chocolate chips, so I threw some of those in between the layers. I also had a few Peppermint Jo Jos on hand so I crushed them up and threw some of those in along with some crushed meringue cookies because, why not? It’s a total smorgasbord of minty chocolate chippy cookie crunchy goodness. I didn’t measure how much of each topping I used, but after I tasted the finished product, I wish I had added more. You can add anything you want- crushed candy canes, peppermint bark, etc. And don’t be afraid to use a heavy hand. Texture = good.

            If you don’t have 7-inch cake pans, you can get 3 shorter 8-inch layers out of this
        }%
        \\
        \\
         \hline
         \parbox{\textwidth}{%
            \begin{center}
                \textbf{Completion (no repetition penalty)}
            \end{center}
            recipe. You can also use a 1-inch layer of chocolate chips, and you can also use a 1-inch layer of chocolate chips.

            Now, let’s talk about the texture. The texture of the cake is the amount of chocolate chips that you’re using. You can use a 1-inch layer of chocolate chips, 1-inch layer of chocolate chips, and 1-inch layer of chocolate chips. You can also use a 1-inch layer of chocolate chips, 1-inch layer of chocolate chips, and 1-inch layer of chocolate chips.

            To make the cake, you’ll need to mix 1-inch chocolate chips, 1-inch chocolate chips, 1-inch chocolate chips, and 1-inch chocolate chips. You’ll also need to mix 1-inch chocolate chips, 1-inch chocolate chips, and 1-inch chocolate chips.
            
            To make the cake, you’ll need to mix 1-inch chocolate chips, 1-inch chocolate chips, and 1-inch chocolate chips. You’ll also need to mix 1-inch chocolate chips, 1-inch chocolate chips, and 1-inch chocolate chips.
        }%
        \\
        \\
        \hline
        \parbox{\textwidth}{%
            \begin{center}
                \textbf{Completion (1.5 repetition penalty)}
            \end{center}
            recipe. The best part is that they are made from 120\% natural ingredients. So if you need to make your own, just go ahead and buy one!

            Now let’s talk about what makes these cakes so special. First, there are two main components: the base layer (the top) and the outer layer (the bottom). This layer consists of three layers: the base layer (which contains all the other layers), the inner layer (which includes the outer layer), and the middle layer (which has everything else). Each layer provides its unique texture and flavor profile. For example, when making a macaroon cake, we would use a thinner layer called the “middle layer” which helps hold the shape together. But if we wanted to create a thicker version, we could choose a lighter layer like a “cream cheese” or even a “dry cream cheese”.
            
            Next, let’s discuss the “inferior layer”. This is where our final dessert comes in handy.
        }%
        \\
        \\
        \hline
    \end{tabular}
\end{table*}

\clearpage

\section{Varying MNTP ratio results on other datasets}
\label{app:vary}

\begin{figure}
    \centering
    \includegraphics[width=0.7\linewidth]{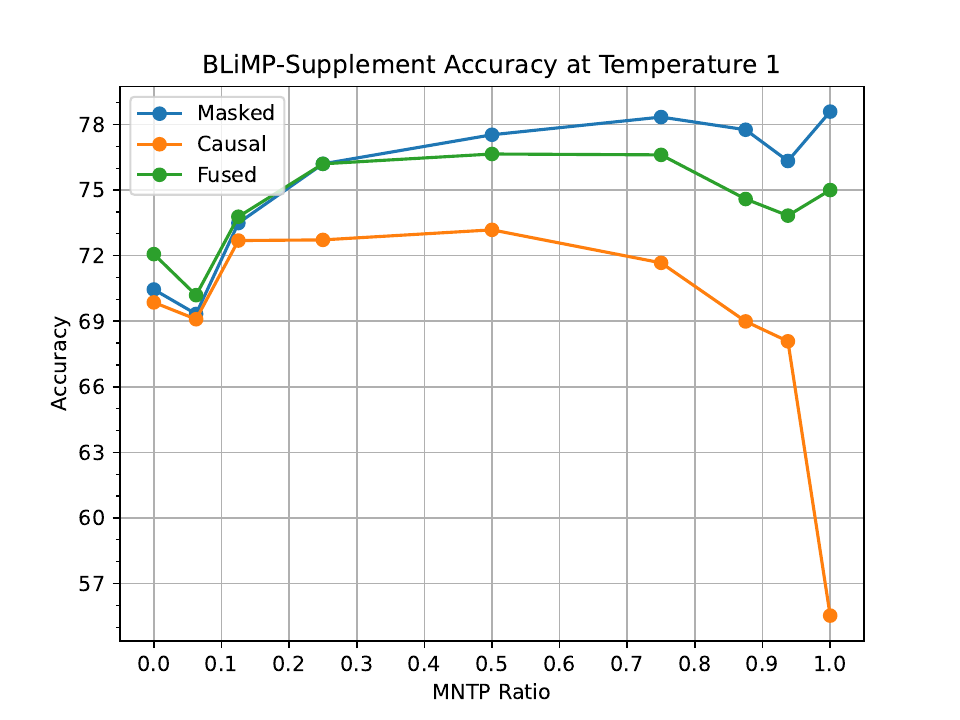}
    \caption{\textbf{BLiMP-Supplement Accuracy}\hspace{1.5em}Comparison of BLiMP-Supplement accuracy when varying the ratio of MNTP used during pre-training. We set the temperature to apply on the logits to 1 for fair comparison between the evaluation strategies. Fused is the sum of the logits from the causal and masked evaluation.}
    \label{fig:blimpsupp1}
\end{figure}

\begin{figure}
    \centering
    \includegraphics[width=0.7\linewidth]{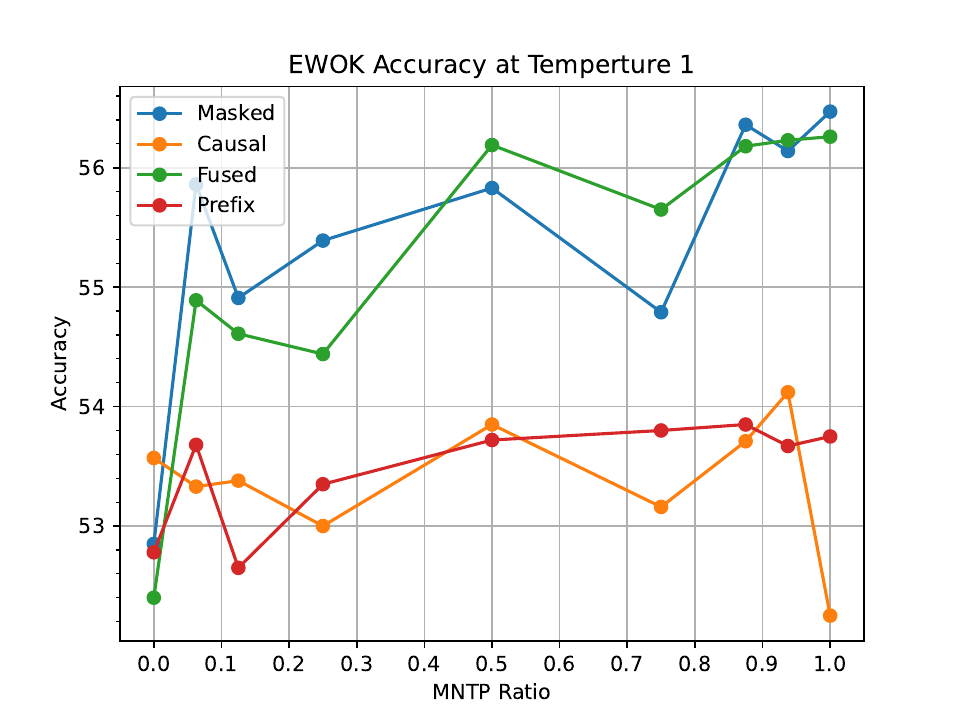}
    \caption{\textbf{EWoK Accuracy}\hspace{1.5em}Comparison of EWoK accuracy when varying the ratio of MNTP used during pre-training. We set the temperature to apply on the logits to 1 for fair comparison between the evaluation strategies. Fused is the sum of the logits from the causal and masked evaluation. We also look at the performance of the model using a prefix masking strategy where the whole context is visible to the model.}
    \label{fig:ewok1}
\end{figure}

\Cref{fig:blimpsupp1,fig:ewok1} show the result of varying the MNTP ratio on the BLiMP-Supplement and EWoK benchmarks. We evaluate the benchmarks with the masked, causal, fused (the sum of the logits of the masked and causal scheme), and prefix (for EWoK) schemes.

\newpage

\section{BLiMP}
\label{app:blimp}

The BabyLM challenge uses the BLiMP benchmark \citep{warstadt-etal-2020-blimp-benchmark} to evaluate the syntactic understanding of the models. 
Our detailed results can be found in \cref{blimp}. 
The component tasks are as follows (with descriptions from \citet{warstadt-etal-2020-blimp-benchmark}):
\begin{description}
    \item[Anaphor Agreement] (AA): the requirement that reflexive pronouns like \textit{herself} (also known as anaphora) agree with their antecedents in person, number, gender, and animacy.
    \item[Argument structure] (AS): the ability of different verbs to appear with different types of arguments. For instance, different verbs can appear with a direct object, participate in the causative alternation, or take an inanimate argument.
    \item[Binding] (B): the structural relationship between a pronoun and its antecedent.
    \item[Control/raising] (CR): syntactic and semantic differences between various types of predicates that embed an infinitival VP. This includes control, raising, and \textit{tough}-movement predicates.
    \item[Determiner-noun agreement] (DNA): number agreement between demonstrative determiners (e.g., \textit{this/these}) and the associated noun.
    \item[Ellipsis] (E): the possibility of omitting expressions from a sentence. Because this is difficult to illustrate with sentences of equal length, our paradigms cover only special cases of noun phrase ellipsis that meet this constraint.
    \item[Filler-gap] (FG): dependencies arising from phrasal movement in, for example, \textit{wh}-questions.
    \item[Irregular forms] (IF): irregular morphology on English past participles (e.g., \textit{awoken}).
    \item[Island effects] (IE): restrictions on syntactic environments where the gap in a filler-gap dependency may occur.
    \item[NPI licensing] (NL): restrictions on the distribution of \textit{negative polarity items} like \textit{any} and \textit{ever} limited to, for example, the scope of negation and \textit{only}.
    \item[Quantifiers] (Q): restrictions on the distribution of quantifiers.  Two such restrictions are covered: superlative quantifiers (e.g., \textit{at least}) cannot be embedded under negation, and definite quantifiers and determiners cannot be subjects in existential-\textit{there} constructions.
    \item[Subject-verb agreement] (SVA): subjects and present tense verbs must agree in number.
\end{description}

On temperature scaling, we observe that for the masked scheme, the increase in performance when using temperature scaling is on average of 2\%. This is not the case for the causal scheme, where temperature seems to have very little effect on the performance of the model.

\begin{table*}[ht!]
\centering
\small
\begin{tabular}{lcccccccccccc@{}}
\toprule
\textbf{Model} & \textbf{AA} & \textbf{AS} & \textbf{B} & \textbf{CR} & \textbf{DNA} & \textbf{E} & \textbf{FG} & \textbf{IF} & \textbf{IE} & \textbf{NL} & \textbf{Q} & \textbf{SVA} \\
\midrule
\multicolumn{5}{@{}l}{\scriptsize{\textsc{strict-small}}}    \\
ELC-BERT*\textsubscript{(2023)} & 89.5 & 72.5 & 68.1 & 72.6 & 93.4 & \textbf{87.4} & \textbf{80.6} & 91.0 & 67.9 & 79.4 & \textbf{75.2} & 88.7 \\[0.5em]
GPT-BERT & \textbf{93.6} & \textbf{78.2} & \textbf{68.8} & \textbf{77.4} & \textbf{97.3} & 86.1 & 80.5 & \textbf{91.5} & \textbf{69.8} & \textbf{84.1} & 68.4 & \textbf{92.2} \\[1em]
\multicolumn{5}{@{}l}{\scriptsize{\textsc{strict}}}    \\
ELC-BERT*\textsubscript{(2023)} & 92.8 & 81.2 & 74.0 & 79.2 & 96.0 & \textbf{91.7} & 87.1 & 93.6 & \textbf{83.9} & 83.5 & 70.2 & 90.8 \\
LTG-BERT*\textsubscript{(2023)} & 96.1 & 79.5 & \textbf{77.1} & 80.3 & 95.4 & \textbf{91.7} & 87.8 & 94.5 & 79.8 & 84.4 & 72.2 & 91.2 \\[0.5em]
GPT-BERT & \textbf{97.7} & \textbf{84.3} & 74.61 & \textbf{83.7} & \textbf{98.2} & 86.9 & \textbf{89.3} & \textbf{96.6} & 77.3 & \textbf{85.2} & \textbf{76.4} & \textbf{95.1} \\
\bottomrule
\end{tabular}
\caption{\label{blimp}
Detailed BLiMP results for models trained on both tracks. The \textbf{bold} results represent the best model for the task. The metric used to measure is accuracy. The results are in percentage. *Results from \citep{georges-gabriel-charpentier-samuel-2023-layers}; they are not directly comparable due to the differences in data filtration between the models as well as the optimized BLiMP temperature being used instead of a general one.
}
\end{table*}

\section{BLiMP Supplemental}

The BLiMP Supplemental was introduced in the last version of the BabyLM Challenge \citep{warstadt-etal-2023-findings}. As for BLiMP it tests the syntactic understanding of models. It consists of the following 5 sub-tasks:
\begin{description}
    \item[Hypernym] Checks whether a word is a superset/subset of another word (for example a dog is a mammal so having a dog means having a mammal).
    \item[QA Congruence Easy] Checks where the question type is congruent with the answer (i.e. a who question answers about a person and not a thing).
    \item[QA Congruence Tricky] Same as before but with more ambiguous cases.
    \item[Subject Aux Inversion] Checking whether the verb relates to the correct subject.
    \item[Turn Talking] Checks whether the right personal pronoun is used in the answer to a question in a conversation.
\end{description}
The results can be found in \cref{suppl}.

\begin{table*}[ht!]
    \centering
    \small
    \begin{tabular}{lcccccccccccc@{}}
        \toprule
        \textbf{Model} & \textbf{Hypernym} & \textbf{QA Cong. Easy} & \textbf{QA Cong. Tricky} & \textbf{Subject Aux Inversion} & \textbf{Turn Talking}\\
        \midrule
        \multicolumn{5}{@{}l}{\scriptsize{\textsc{strict-small}}}    \\
        Encoder\textsubscript{\textit{(baseline)}} & \textbf{54.2} & 62.5 & 49.1 & 79.9 & 58.2 \\
        Decoder\textsubscript{\textit{(baseline)}} & 49.6 & 54.7 & 41.2 & 86.0 & 66.1 \\[0.5em]
        ELC-BERT*\textsubscript{\textit{(2023)}}   & 48.0 & \textbf{73.4} & 43.6 & \textbf{90.0} & 84.3 \\[0.5em]
        GPT-BERT                                    & 47.1 & \textbf{73.4} & \textbf{54.5} & 86.3 & \textbf{85.7} \\[1em]
        \multicolumn{5}{@{}l}{\scriptsize{\textsc{strict}}}    \\
        Encoder\textsubscript{\textit{(baseline)}} & \textbf{55.0} & 75.0 & 53.3 & 87.5 & 61.4 \\
        Decoder\textsubscript{\textit{(baseline)}} & 45.6 & 56.2 & 44.8 & 83.9 & 72.5 \\[0.5em]
        ELC-BERT*\textsubscript{\textit{(2023)}}   & 47.3 & 85.9 & \textbf{63.0} & 94.5 & \textbf{92.1} \\
        LTG-BERT*\textsubscript{\textit{(2023)}}   & 47.0 & \textbf{90.6} & 60.6 & 90.7 & \textbf{92.1} \\[0.5em]
        GPT-BERT                                    & 48.8 & \textbf{90.6} & 59.4 & \textbf{96.3} & 88.9 \\
        \bottomrule
    \end{tabular}
    \caption{\label{suppl}
    Detailed BLiMP supplemental results for models trained on both tracks. The \textbf{bold} results represent the best model for the task. The metric used to measure performance is accuracy. The results are in percentage. *Results from \citep{georges-gabriel-charpentier-samuel-2023-layers}; they are not directly comparable due to the differences in data filtration between the models as well as the optimized BLiMP Supplemental temperature being used instead of a general one.
    }
\end{table*}

\section{GLUE}
\label{app:glue}

The BabyLM challenge involves slightly modified GLUE and SuperGLUE benchmarks. It uses only a subset of the subtasks, the datasets are filtered so that they do not contain out-of-vocabulary words, and it sometimes uses non-standard metrics. 
Our detailed results can be found in \cref{glue}. 
We list all subtasks and their metrics below:

\begin{description}\itemsep0em 
    \item[Boolean Questions] \citep[BoolQ;][]{clark-etal-2019-boolq}, a yes/no Q/A dataset evaluated with accuracy.
    \item[Corpus of Linguistic Acceptability] \citep[CoLA;][]{warstadt-etal-2019-neural} evaluated with the Matthews correlation coefficient \citep[MCC;][]{MATTHEWS1975442}.
    \item[The Multi-Genre Natural Language Inference Corpus] \citep[MNLI;][]{williams-etal-2018-broad}. Its development set consists of two parts: \textit{matched}, sampled from the same data source as the training set, and \textit{mismatched}, which is sampled from a different domain. Both parts are evaluated with accuracy.
    \item[The Microsoft Research Paraphrase Corpus] \citep[MRPC;][]{dolan-brockett-2005-automatically}, evaluated with both F\textsubscript{1}-score (originally also evaluated with accuracy).
    \item[Multi-Sentence Reading Comprehension] \citep[MultiRC;][]{khashabi-etal-2018-looking}, a multiple choice question answering dataset, evaluated with accuracy (originally evaluated with the exact match accuracy (EM) and F\textsubscript{1}-score (over all answer options)).
    \item[Question-answering Natural Language Inference] (QNLI) constructed from the Stanford Question Answering Dataset \citep[SQuAD;][]{rajpurkar-etal-2016-squad}, evaluated with accuracy.
    \item[The Quora Question Pairs] (QQP),\footnote{\url{https://quoradata.quora.com/First-Quora-Dataset-Release-Question-Pairs}} evaluated with F\textsubscript{1}-score (originally evaluated with accuracy).
    \item[The Stanford Sentiment Treebank] \citep[SST-2;][]{socher-etal-2013-recursive}, evaluated with accuracy.
    \item[The Recognizing Textual Entailment datasets] \citep[RTE;][]{10.1007/11736790_9, rte2, giampiccolo-etal-2007-third, Bentivogli09thefifth}, evaluated with accuracy.
    \item[Winograd Schema Challenge] \citep[WSC;][]{10.5555/3031843.3031909} evaluated with accuracy.
\end{description}

\begin{table}[ht!]
\resizebox{\textwidth}{!}{
\begin{tabular}{lccccccccccc@{}}
\toprule
\textbf{Model} & \textbf{CoLA} & \textbf{SST-2} & \textbf{MRPC} & \textbf{QQP} & \textbf{MNLI\textsubscript{m}} & \textbf{MNLI\textsubscript{mm}} & \textbf{QNLI} & \textbf{RTE} & \textbf{BoolQ} & \textbf{MultiRC} & \textbf{WSC}\\

\midrule
\multicolumn{5}{@{}l}{\small{\textsc{strict-small}}}    \\

Encoder\textsubscript{\textit{(baseline)}} & 0.0 & 85.1 & 82.2 & 34.2 & 68.9 & 68.9 & 76.5 & 58.3 & 68.8 & 58.5 & 61.5 \\

Decoder\textsubscript{\textit{(baseline)}} & 2.2 & 86.2 & 82.0 & 83.6 & 72.4 & 74.2 & 82.8 & 49.6 & 65.0 & 60.1 & 38.5 \\[0.5em]

ELC-BERT*\textsubscript{\textit{(2023)}} & -- & 89.3$^{\pm 0.5}$ & 85.0$^{\pm 1.8}$ & 86.7$^{\pm 0.3}$ & 79.2$^{\pm 0.3}$ & 79.9$^{\pm 0.2}$ & 85.8$^{\pm 0.4}$ & 55.4$^{\pm 2.6}$ & 69.3$^{\pm 2.0}$ & 62.2$^{\pm 1.0}$ & 59.0$^{\pm 5.4}$ \\

LTG-BERT*\textsubscript{\textit{(2023)}} & -- & 88.8$^{\pm 0.8}$ & 82.3$^{\pm 0.4}$ & 85.8$^{\pm 0.2}$ & 78.0$^{\pm 0.2}$ & 78.8$^{\pm 0.4}$ & 85.0$^{\pm 0.2}$ & 53.7$^{\pm 4.1}$ & 64.8$^{\pm 2.1}$ & 64.1$^{\pm 0.3}$ & 60.5$^{\pm 1.0}$ \\[0.5em]

GPT-BERT & \textbf{48.9} & \textbf{92.2} & \textbf{91.5} & \textbf{87.1} & \textbf{80.2} & \textbf{80.5} & \textbf{86.4} & \textbf{64.0} & \textbf{72.5} & \textbf{69.3} & \textbf{69.2} \\[1em]

\multicolumn{5}{@{}l}{\small{\textsc{strict}}}    \\

Encoder\textsubscript{\textit{(baseline)}} & 34.6 & 91.5 & 83.1 & 86.7 & 77.7 & 78.1 & 78.2 & 46.8 & 61.7 & 52.6 & 61.5 \\

Decoder\textsubscript{\textit{(baseline)}} & 37.3 & 88.3 & 86.8 & 84.5 & 75.6 & 76.2 & 83.1 & 60.4 & 66.1 & 62.1 & 38.5 \\[0.5em]

ELC-BERT*\textsubscript{(2023)} & -- & 91.9$^{\pm 1.1}$ & 89.3$^{\pm 0.6}$ & 88.0$^{\pm 0.1}$ & 83.6$^{\pm 0.1}$ & 83.3$^{\pm 0.2}$ & 89.4$^{\pm 0.4}$ & 60.0$^{\pm 2.8}$ & 70.5$^{\pm 1.5}$ & 66.2$^{\pm 2.2}$ & 56.4$^{\pm 9.4}$ \\

LTG-BERT*\textsubscript{(2023)} & -- & 92.0$^{\pm 0.4}$ & 87.4$^{\pm 0.7}$ & 87.9$^{\pm 0.1}$ & 83.0$^{\pm 0.4}$ & 83.4$^{\pm 0.5}$ & 89.1$^{\pm 0.5}$ & 54.7$^{\pm 2.4}$ & 68.4$^{\pm 0.5}$ & 66.0$^{\pm 1.4}$ & 61.4$^{\pm 0.0}$ \\[0.5em]

GPT-BERT & \textbf{62.4} & \textbf{94.0} & \textbf{94.4} & \textbf{89.1} & \textbf{85.2} & \textbf{85.3} & \textbf{90.8} & \textbf{69.1} & \textbf{78.4} & \textbf{73.3} & \textbf{75.0} \\

\bottomrule
\end{tabular}
}
\caption{\label{glue}
A subset of GLUE results (defined by the Baby LM challenge) for models trained on both tracks. All the results indicate the model accuracy for the task except for MRPC and QQP where the results are based on the F1-score of the positive class and CoLA which reports the MCC. The results are reported in percentage. The \textbf{bold} result indicates the best model for each dataset. *Results from \citep{georges-gabriel-charpentier-samuel-2023-layers}; they are not directly comparable due to the differences in data filtration between the models.
}
\end{table}

\section{EWoK}
\label{app:ewok}

The BabyLM challenge uses a slightly modified EWoK benchmark \citep{ivanova2024elementsworldknowledgeewok}. It tests all concepts but filters the dataset to include only examples where the words appear in the BabyLM dataset. 
Our detailed results can be found in \cref{tab:ewok}. 
We list all concepts below:

\begin{description}\itemsep0em 
    \item[Agent Properties] Checks whether the model can recognize agent (conscious beings) properties (such as believe, choice, feeling, etc.)
    \item[Material Dynamics] Checks whether the model can recognize the dynamics (movement, fluidity, etc.) of a given material.
    \item[Material Properties] Checks whether the model can recognize the properties (bounciness, hardness, etc.) of a given material.
    \item[Physical Dynamics] Checks whether the model can recognize the physical dynamic (speed, buoyancy, etc.) of an object.
    \item[Physical Interactions] Checks whether the model can recognize the physical interactions (attraction, collision, etc.) between objects.
    \item[Physical Relations] Checks whether the model can recognize the physical relations (attached vs. connected, bigger vs. smaller, etc.) between objects.
    \item[Quantative Properties] Checks whether the model can recognize amount (a lot vs. little of, enough vs. not enough, etc.) of an object.
    \item[Social Interactions] Checks whether the model can recognize the social interactions (cooperate vs. compete, help vs. deceive, etc.) between agents.
    \item[Social Properties] Checks whether the model can recognize the social property (boastful vs. humble, dominant vs. submissive, etc.) of an agent.
    \item[Social Relations] Checks whether the model can recognize the social relations (boss vs. subordinate, colleague vs. boss, etc.) between agents.
    \item[Spatial Relations] Checks whether the model can recognize the spatial relations (location, height, etc.) between agents, objects or a combination of them.
\end{description}

\begin{table}[]
    \small
    \centering
    \begin{tabular}{lccccccccccc@{}}
        \toprule
        \multirow{2}{8em}{\textbf{Model}} & \textbf{Agent} & \multicolumn{2}{c}{\textbf{Material}} & \multicolumn{3}{c}{\textbf{Physical}} & \textbf{Quantative} & \multicolumn{3}{c}{\textbf{Social}} & \textbf{Spatial} \\
        \cmidrule(lr){3-4} \cmidrule(lr){5-7} \cmidrule(lr){9-11}
        & \textbf{Prop.} & \textbf{Dyn.} & \textbf{Prop.} & \textbf{Dyn.} & \textbf{Inter.} & \textbf{Rel.} & \textbf{Prop.} & \textbf{Inter.} & \textbf{Prop.} & \textbf{Rel.} & \textbf{Rel.} \\
        \midrule
        \multicolumn{2}{@{}l}{\scriptsize\textsc{strict-small}}\\
        Encoder\textsubscript{\textit{(baseline)}} & 50.2 & 51.0 & 45.3 & 42.5 & 49.1 & \textbf{51.0} & 48.1 & 51.7 & 53.4 & 50.6 & 45.3 \\
        Decoder\textsubscript{\textit{(baseline)}} & 50.5 & 51.7 & \textbf{49.4} & 54.2 & 50.4 & 50.6 & 53.5 & 50.7 & 50.3 & 49.8 & 46.7 \\[0.5em]
        GPT-BERT                                    & \textbf{50.7} & \textbf{58.1} & 48.8 & \textbf{57.5} & \textbf{51.1} & 49.9 & \textbf{55.7} & \textbf{65.6} & \textbf{58.2} & \textbf{51.6} & \textbf{53.9} \\[1em]
        \multicolumn{2}{@{}l}{\scriptsize\textsc{strict}}\\
        Encoder\textsubscript{\textit{(baseline)}} & 50.1 & 55.8 & 50.6 & \textbf{58.3} & 48.9 & \textbf{50.9} & 53.8 & 51.4 & 50.8 & 53.8 & 51.4 \\
        Decoder\textsubscript{\textit{(baseline)}} & 50.1 & 55.5 & 50.0 & 57.5 & 51.4 & 50.5 & \textbf{56.7} & 52.7 & 49.7 & 50.0 & 49.0 \\[0.5em]
        GPT-BERT                                    & \textbf{52.7} & \textbf{72.3} & \textbf{51.8} & 50.8 & \textbf{52.7} & 48.3 & 52.5 & \textbf{77.2} & \textbf{64.3} & \textbf{58.9} & \textbf{60.8} \\
        \bottomrule
    \end{tabular}
    \caption{Detailed EWoK results for models trained on both tracks. The \textbf{bold} results represent the best model for the task. The metric used to measure performance is accuracy.}
    \label{tab:ewok}
\end{table}

\section{LAMBADA}
\label{sec:lambada}

LAMBADA is a zero-shot language modeling task that focuses on resolving long-range dependencies in text \citep{paperno-etal-2016-lambada}; we used its detokenized version from \newcite{radford2019language}. While it has been traditionally used for evaluating autoregressive language models, we adapt the task for masked language models. 
Note that this adaptation does not allow for a direct comparison with the autoregressive models. An illustrative sample from this dataset looks as follows:

\textbf{Prompt:} \textit{"Give me a minute to change and I'll meet you at the docks." She'd forced those words through her teeth. "No need to change. We won't be that long." Shane gripped her arm and started leading her to the dock. "I can make it there on my own, \textbf{\{answer\}}."}

\textbf{Gold answer:} \textit{Shane}

We insert the whole tokenized prompt to the evaluated language model and replace the missing answer by $k$ mask tokens, where $k$ is the length of the tokenized gold answer. Then we evaluate the exact-match accuracy of predicting filling in the correct continuation and also the mean perplexity. 

We also evaluate using the normal causal method implemented by \newcite{radford2019language}, as well as doing it with a prefix, where all the context tokens attend to each other.

\end{document}